\documentclass[a4paper]{cas-sc}
\usepackage[authoryear]{natbib}
\setcitestyle{authoryear,round,aysep={,},citesep={;}}

\usepackage{graphicx}
\usepackage{amsmath,amssymb}
\usepackage{booktabs,multirow}
\usepackage{etoolbox,xspace,url}
\usepackage{float}
\usepackage[section]{placeins}
\usepackage{multirow}
\usepackage{array}
\def\tsc#1{\csdef{#1}{\textsc{\lowercase{#1}}\xspace}}
\tsc{WGM}
\tsc{QE}
\usepackage{xcolor}
\definecolor{heppurple}{HTML}{E44EE4}

\usepackage{caption}

\DeclareCaptionLabelSeparator{myperiod}{. }
\usepackage{amsmath,amssymb}
\usepackage{mathtools}

\AtBeginDocument{
  \setlength{\abovedisplayskip}{3pt plus 1pt minus 1pt}
  \setlength{\belowdisplayskip}{3pt plus 1pt minus 1pt}
  \setlength{\abovedisplayshortskip}{1pt plus 1pt minus 1pt}
  \setlength{\belowdisplayshortskip}{3pt plus 1pt minus 1pt}
}

\usepackage{amsmath}
\usepackage{bm}
\usepackage{mathtools}
\usepackage{caption}
\usepackage[section]{placeins} % Prevent figures/tables from drifting beyond the next section.
\usepackage{algorithm}
\usepackage{algpseudocode}
\usepackage{float}
\def\tsc#1{\csdef{#1}{\textsc{\lowercase{#1}}\xspace}}
\tsc{HEP}
\tsc{MRI}
\begin{document}
\let\WriteBookmarks\relax
%------------------------- 页眉短标题 / 作者缩写 -------------------------
\shorttitle{High-dimensional Embedding Prior for Noisy K-space Domain MRI Reconstruction}
\shortauthors{Guan Y, Huang TJ, Cai QR, et al.}

%------------------------- 论文主标题 -------------------------
\title[mode = title]{High-dimensional Embedding Prior for Noisy K-space Domain MRI Reconstruction}
\tnotemark[1]
\tnotetext[1]{This work was supported by the National Key Research and Development Program of China, Grant Nos: 2023YFF1204300, 2023YFF1204302.}

%------------------------- 作者信息（零报错格式） -------------------------
\author[1,3]{Yu Guan}
\fnmark[1]
\ead{guanyu@ncu.edu.cn}

\author[2,3]{Tianjia Huang}
\fnmark[1]
\ead{huangtianjia@ncu.edu.cn}

\author[3]{Qinrong Cai}
\ead{caiqinrong@ncu.edu.cn}

\author[4]{Qiuyun Fan}
\ead{fanqiuyun@tju.edu.cn}

\author[5]{Dong Liang}
\ead{liangdong@siat.ac.cn}

\author[3]{Qiegen Liu}
\cormark[1]
\ead{liuqiegen@ncu.edu.cn}
\ead[url]{https://www.ncu.edu.cn}

%------------------------- 作者贡献（CRediT） -------------------------
\credit{Conceptualization, Methodology, Software, Writing - original draft}
\credit{Data curation, Validation, Formal analysis}
\credit{Supervision, Project administration, Funding acquisition, Writing - review & editing}

%------------------------- 单位信息 -------------------------
\affiliation[1]{organization={School of Advanced Manufacturing, Nanchang University},
    city={Nanchang},
    postcode={330031},
    country={China}}

\affiliation[2]{organization={School of Mathematics and Computer Science, Nanchang University},
    city={Nanchang},
    postcode={330031},
    country={China}}

\affiliation[3]{organization={School of Information Engineering, Nanchang University},
    city={Nanchang},
    postcode={330031},
    country={China}}

\affiliation[4]{organization={Academy of Medical Engineering and Translational Medicine, Medical School, Faculty of Medicine, Tianjin University},
    city={Tianjin},
    postcode={300072},
    country={China}}

\affiliation[5]{organization={Shenzhen Institute of Advanced Technology, Chinese Academy of Sciences},
    city={Shenzhen},
    postcode={518000},
    country={China}}

%------------------------- 脚注 -------------------------
\fntext[1]{These authors contributed equally to this work.}
\cortext[1]{Corresponding author: Qiegen Liu, liuqiegen@ncu.edu.cn}

\nonumnote{}

\begin{abstract}
Magnetic resonance imaging (MRI) reconstruction under realistic acquisition conditions can be fundamentally viewed as estimating the underlying k-space distribution from incomplete and noise-corrupted measurements. While diffusion models have recently shown strong potential as generative prior for inverse problems, existing approaches struggle to handle noisy reconstruction settings, especially when operating directly in k-space domain. In this work, we propose a unified high-dimensional k-space reconstruction framework tailored for noisy inverse problems, which enhances diffusion-based solvers through representation lifting. Rather than modifying the underlying optimization procedures, the proposed framework augments the data representation space, enabling existing diffusion-based solvers to operate on enriched k-space embeddings with improved expressiveness. Extensive experiments on both in-house and public datasets across varying noise levels and undersampled factors demonstrate that the proposed framework consistently improves reconstruction quality for multiple diffusion-based inverse solvers. Notably, the largest gains are observed in high-noise regimes, which is consistent with our theoretical analysis of error propagation under high-dimensional representation. These results suggest that high-dimensional representation provides a general and model-agnostic mechanism for improving diffusion-based MRI reconstruction in noisy settings, offering a new perspective on robust k-space generative modeling for practical inverse problems. The code will be available at \url{https://github.com/yqx7150/HEP-MRIRec}.
\end{abstract}

% Use only if graphical abstract is required by the submission system.
%\includegraphics[width=\linewidth]{figs/graphical_abstract.pdf}

\begin{graphicalabstract}
% 可统一控制图宽，避免图片撑满或过小，按需修改 0.95\textwidth
\includegraphics[width=0.95\textwidth]{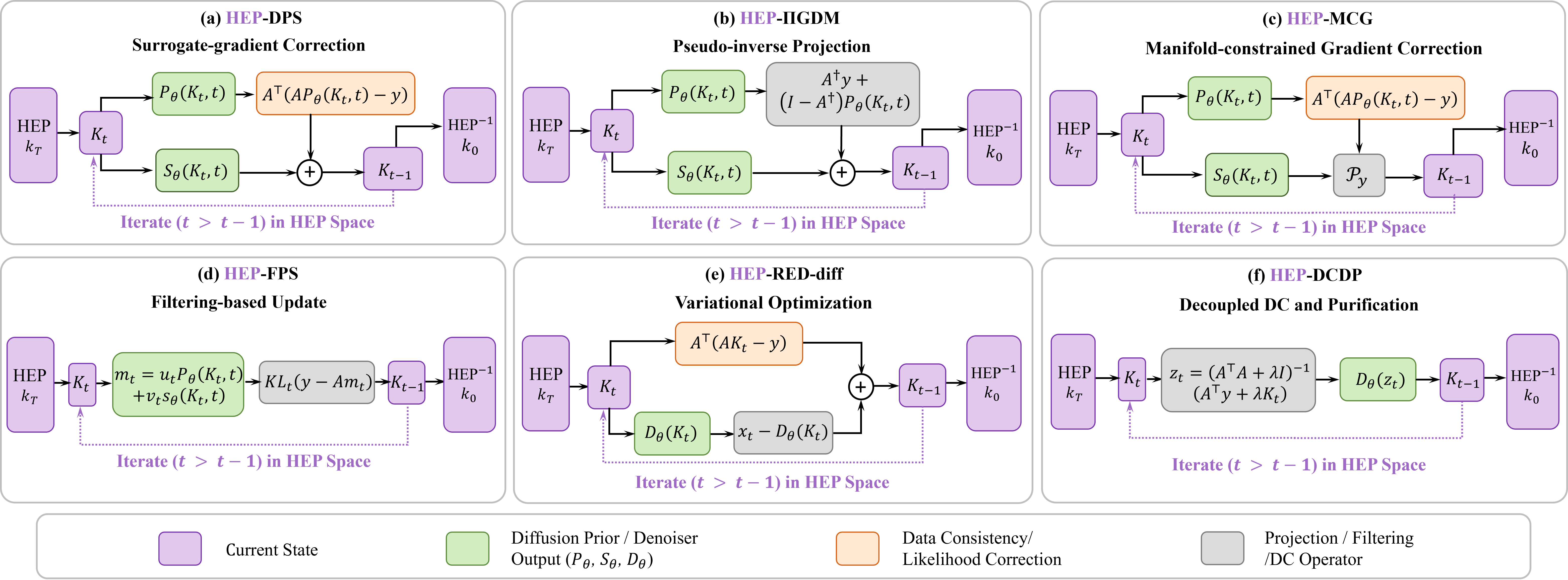}
\end{graphicalabstract}

\begin{highlights}
\item We identify k-space representation as a critical factor in diffusion-based MRI reconstruction and propose a unified HEP framework that elevates k-space data into a more expressive representation space. Through a single embedding operator, HEP provides a plug-and-play mechanism for integrating high-dimensional prior into existing diffusion-based inverse solvers, allowing diverse diffusion-based algorithms to benefit from enriched k-space representation without altering their underlying optimization procedures.
\item We provide a theoretical analysis of error propagation in diffusion-based reconstruction, revealing how representation-induced errors influence score estimation and interact with solver-specific operators during iterative sampling. This analysis offers a principled explanation for the performance variations across different inverse solvers and clarifies the role of HEP in improving reconstruction stability.
\item We conduct extensive experiments under diverse degradation settings, including varying noise levels, undersampling factors and sampling trajectories, across both in-house and public datasets. Results demonstrate consistent improvements across multiple diffusion-based inverse solvers, particularly in noisy and highly ill-posed regimes. We further demonstrate that HEP is complementary to existing prior, such as low-rank constraints, highlighting its flexibility and general applicability.
\end{highlights}


\maketitle

\section{Introduction}\label{sec:introduction}

Magnetic resonance imaging (MRI) is a widely used non-invasive imaging modality in clinical practice due to its superior soft-tissue contrast and high spatial resolution \citep{Lustig2007,Knoll2020}. However, MRI acquisition remains inherently slow because k-space data must be sequentially sampled, which limits clinical throughput and may introduce motion artifacts and patient discomfort. To address these limitations, accelerated MRI aims to recover high-fidelity MR signals from undersampled k-space measurements, which can be formulated as an ill-posed inverse problem in the frequency domain \citep{Lustig2007}. In practical scenarios, this problem is further complicated by measurement noise, making k-space reconstruction a challenging noisy inverse problem. 

Classical reconstruction methods address the inverse problem by incorporating hand-crafted prior to regularize the solution space. For example, compressed sensing exploits sparsity in predefined transform domains \citep{Lustig2007}, while parallel imaging methods such as SENSE and GRAPPA leverage coil sensitivity information to recover missing k-space data \citep{Pruessmann1999,Griswold2002}. Despite their effectiveness, these methods generally rely on hand-crafted prior, which impose relatively simplified assumptions on the underlying MR signal structure and therefore limit their modeling capacity. This mismatch becomes particularly pronounced under challenging scenarios involving severe undersampling and measurement noise, where the reconstruction problem becomes highly ill-posed and the available measurements are insufficient to determine a unique solution. Consequently, the reconstructed images often suffer from loss of fine details and residual artifacts.

Deep learning has significantly advanced MRI reconstruction by learning data-driven prior directly from large-scale datasets. Early approaches employed convolutional neural networks to learn mappings from undersampled measurements to high-quality reconstructed images \citep{Wang2016,Lee2017,Zbontar2018}.Subsequently, model-based deep learning methods incorporated physical acquisition models into unrolled optimization frameworks, thereby improving reconstruction fidelity and interpretability \citep{Aggarwal2018,Schlemper2017}. However, most existing approaches are still trained as deterministic regression models, optimizing for point-wise reconstruction accuracy rather than explicitly modeling the distribution of MRI data. This limitation results in insufficient uncertainty modeling, making these methods prone to averaging effects when measurements are ambiguous. As a result, they often produce overly smooth reconstructions in highly ill-posed settings.

To overcome these limitations, generative modeling approaches have recently been introduced for inverse problems. In particular, diffusion models provide a principled framework for modeling high-dimensional data distributions through a progressive denoising process \citep{Ho2020,Song2021a}. By learning either the noise or the score function (i.e., the gradient of the log-density), diffusion models define an implicit data prior that can be naturally integrated into inverse problem formulations. This enables posterior sampling by combining learned prior with measurement constraints, making diffusion models well-suited for ill-posed problems such as MRI reconstruction. Recent works have successfully adapted diffusion models to MRI reconstruction by incorporating data consistency into the reverse process, including diffusion posterior sampling \citep{Chung2023}, manifold-constrained guidance \citep{Chung2022} and adaptive diffusion prior \citep{Gungor2023}, demonstrating strong reconstruction performance under undersampled and noisy conditions.

Despite the remarkable progress of diffusion-based MRI reconstruction, existing methods have primarily focused on posterior sampling strategies, data-consistency enforcement and solver-specific update mechanisms \citep{Chung2025,Daras2024}. Comparatively less attention has been paid to the representation of the reconstruction variable itself. This limitation is particularly important for direct MRI reconstruction under realistic acquisition conditions, which can be fundamentally viewed as a noisy k-space signal estimation problem. Due to the measured k-space is simultaneously degraded by undersampling and noise, requiring the reconstruction algorithm to recover missing frequency components while accurately characterizing the underlying signal structure \citep{Cole2020,Singh2022,Ryu2021}. Nevertheless, most current methods represent k-space signal using relatively simple low-dimensional forms, such as real-imaginary channel concatenation, and largely assume that posterior inference can be improved through better sampling or optimization strategies alone. Such representations may be insufficient to capture the complex statistical structure of k-space signals under severe degradation, potentially limiting the quality of score estimation and subsequent posterior inference. As a result, a systematic understanding of how signal representation influences diffusion prior learning and posterior inference in noisy k-space reconstruction remains largely unexplored.

Previous research from our group have provided initial evidence that representation design can play an important role in diffusion-based MRI reconstruction. In particular, weighted k-space modeling \citep{Tu2023} demonstrated that transforming the original k-space representation can improve the effectiveness of diffusion prior and lead to improved reconstruction quality. Following this research line, we adapted the single representation to a dual-mode framework \citep{Guan2024}. By integrating weighted and masked k-space representations, we developed a multi-frequency diffusion framework that exploits cross-frequency correlations to guide the denoising process, resulting in improved reconstruction accuracy and faster convergence. Moreover, we extended this research direction from frequency-aware representations to high-dimensional prior modeling for multi-contrast MRI reconstruction \citep{Guan2025}. Specifically, the proposed framework constructs a subset k-space through distribution matching and embeds high-dimensional global prior into the diffusion process, which enhances prior expressiveness and suppresses reconstruction artifacts. These findings suggest that representation is not merely a data preprocessing step, but a fundamental factor influencing posterior inference. Nevertheless, existing explorations remain task-specific and lack a general theoretical framework capable of explaining how representation transformations affect diffusion-based inverse solvers under realistic noisy acquisition conditions.

Building upon our previous studies on k-space representation learning and diffusion-based MRI reconstruction, we further investigate the inverse problem of noisy k-space reconstruction under realistic acquisition conditions. From the theoretical perspective, we develop a unified error-propagation analysis that characterizes how representation-induced prior errors interact with data-consistency constraints and propagate through different diffusion-based solver architectures. From the algorithmic perspective, we propose a High-dimensional Embedding Prior (HEP) framework that systematically lifts k-space signals into more expressive representation spaces to diffusion inference. Specifically, structured high-dimensional embeddings, including multi-channel and  wavelet transform representations, are introduced to enrich the statistical structure of k-space signals while maintaining compatibility with a broad range of diffusion-based reconstruction methods. By connecting representation learning with noisy inverse problem solving, the proposed framework provides both a principled explanation for the role of high-dimensional representation and a practical strategy for improving diffusion-based MRI reconstruction under realistic acquisition conditions.

The main contributions of this work are summarized as follows:
\begin{itemize}
    \item We identify k-space representation as a critical factor in diffusion-based MRI reconstruction and propose a unified HEP framework that elevates k-space data into a more expressive representation space. Through a single embedding operator, HEP provides a plug-and-play mechanism for integrating high-dimensional prior into existing diffusion-based inverse solvers, allowing diverse diffusion-based algorithms to benefit from enriched k-space representation without altering their underlying optimization procedures.
    
    \item We provide a theoretical analysis of error propagation in diffusion-based reconstruction, revealing how representation-induced errors influence score estimation and interact with solver-specific operators during iterative sampling. This analysis offers a principled explanation for the performance variations across different inverse solvers and clarifies the role of HEP in improving noisy k-space reconstruction stability.
    
    \item We conduct extensive experiments under diverse degradation settings, including varying noise levels, undersampling factors and sampling trajectories, across both in-house and public datasets. Results demonstrate consistent improvements across multiple diffusion-based inverse solvers, particularly in noisy and highly ill-posed regimes. We further demonstrate that HEP is complementary to existing prior, such as low-rank constraints, highlighting its flexibility and general applicability.
\end{itemize}

\section{Preliminaries}\label{sec:preliminaries}

\subsection{Diffusion-based Generative prior}\label{subsec:diffusion-generative-prior}

Diffusion models provide a powerful framework for modeling complex high-dimensional data distributions through a progressive noising and denoising process \citep{SohlDickstein2015,Ho2020}. Starting from clean data samples, the forward process gradually perturbs the data toward a tractable Gaussian distribution, while the reverse process learns to reconstruct the data by removing noise step by step.

Generally, the forward process admits a continuous-time formulation via a stochastic differential equation:
\begin{equation}
\operatorname{d}\!x_t = \left[ f\left(x_t,t\right) - g^2\left(t\right)\nabla_{\!x_t}\log p_t\left(x_t\right) \right] \operatorname{d}\!t + g\left(t\right) \operatorname{d}\!\bar{w}_t,
\label{eq:reverse_sde}
\end{equation}
where $f\left(x_t,t\right)$ denotes the drift coefficient, $g\left(t\right)$ controls the diffusion strength, and $w_t$ is a standard Wiener process. By appropriately choosing $f$ and $g$, the marginal distribution $p_t\left(x_t\right)$ evolves smoothly toward an isotropic Gaussian prior. Common diffusion parameterizations include variance-preserving and variance-exploding forward processes, both of which recover widely used discrete diffusion models as special cases and provide a unified mathematical description of noise perturbation dynamics. \citep{Song2021a,Karras2022}.

Sampling from the data distribution is achieved by simulating the reverse-time dynamics associated with the forward process. Under mild regularity conditions, the reverse-time process exists and is given by
\begin{equation}
\operatorname{d}\!x_t = \left[f\left(x_t,t\right) - g^2\left(t\right)\nabla_{x_t}\log{p_t}\left(x_t\right)\right]\operatorname{d}\!t + g\left(t\right)\operatorname{d}\!\bar{w}_t,
\label{eq:reverse_sde2}
\end{equation}
where $\nabla_{x_t}\log{p_t}\left(x_t\right)$ is the score function of the perturbed data distribution, and $\bar{w}_t$ denotes a reverse-time Wiener process \citep{Vincent2011,Song2019,Song2021a}. Intuitively, the forward process gradually destroys structure by adding Gaussian noise, while the reverse process reconstructs structure by following the gradient of the log-density. In practice, the score function is approximated by a neural network \(s_\theta\left(x_t,t\right)\) trained using conventional denoising score matching \citep{Hyvarinen2005,Song2020}. Once trained, the model provides access to the data prior through score evaluations without requiring explicit likelihood computation.

An important property of diffusion models is their connection to conditional expectation under Gaussian perturbations. Specifically, Tweedie's formula establishes that
\begin{equation}
\mathbb{E}\left[x_0 \mid x_t\right] = x_t + \sigma_t^2\nabla_{x_t}\log{p_t}\left(x_t\right),
\label{eq:tweedie_formula}
\end{equation}
where $x_0$ denotes the clean data sample and $\sigma_t$ represents the noise level at time $t$ \citep{Robbins1956,Efron2011,Kingma2021}. This result reveals that the score function implicitly encodes the estimate of the clean signal, which forms the theoretical foundation for denoising in diffusion models.

For notational consistency across all algorithms, we formally define the clean-signal prediction function as the practical implementation of this theoretical estimate. Empirically, the intractable score function \(\nabla_{x_t}\log{p_t}\left(x_t\right)\) is approximated by a parameterized neural network \(s_\theta\left(x_t,t\right)\) \citep{Vincent2011,Ho2020,Song2021a,Song2021b}. Substituting this learned approximation into Eq.~\eqref{eq:tweedie_formula} yields a computable estimator of the clean signal, which we uniformly denote as \(P_\theta\left(x_t,t\right)\) in all subsequent algorithms.
\begin{equation}
P_\theta\left(x_t,t\right) = x_t + \sigma_t^2s_\theta\left(x_t,t\right).
\label{eq:clean_signal_prediction}
\end{equation}

From this perspective, diffusion models define an implicit prior over the data distribution through two coupled operators: the score function $s_\theta\left(x_t,t\right)$ that characterizes the gradient of the log-density and the clean-signal prediction function $P_\theta\left(x_t,t\right)$ that directly estimates the underlying clean signal. Together, these two operators provide consistent characterizations of the same data manifold and serve as the core building blocks for the diffusion-based inverse problem algorithms presented in the following sections.

\subsection{Diffusion-based Posterior Sampling for Inverse Problems}\label{subsec:diffusion-posterior-sampling}

We now consider inverse problems of the form
\begin{equation}
y = Ax_0 + n,\quad n \sim \mathcal{N}\left(0,I\right),
\label{eq:inverse_problem}
\end{equation}

where $A$ denotes the forward measurement operator, $y$ represents the observed measurements, and $x_0$ is the unknown signal to be recovered under a diffusion-based prior \citep{Meng2025,Boys2024,Rout2024}. In this section, we use a unified notation to describe several representative diffusion-based inverse solvers. Specifically, $s_\theta\left(x_t,t\right)$ denotes the learned score function that drives the reverse diffusion process, $P_\theta\left(x_t,t\right)$ denotes the clean-signal prediction or denoised estimate induced by Tweedie's formula, and $D_\theta\left(\cdot\right)$ denotes a diffusion denoiser used as a regularization or purification operator.

Within the diffusion framework introduced in Section~\ref{subsec:diffusion-generative-prior}, posterior sampling is performed along the reverse-time trajectory. The associated probability flow can be expressed as
\begin{equation}
\frac{\operatorname{d}\!x_t}{\operatorname{d}\!t}
=
f\left(x_t,t\right)
-
g^2\left(t\right)
\left[
s_\theta\left(x_t,t\right)
-
\nabla_{x_t}\log p\left(y\mid x_t\right)
\right],
\label{eq:posterior_flow}
\end{equation}
The first term encodes learned diffusion prior via score function, and the second ensures measurement consistency \citep{Song2022}. The score term is computable via trained diffusion models, yet the likelihood term remains intractable, requiring marginalization over unknown clean signals:
\begin{equation}
\nabla_{x_t}\log p\left(y\mid x_t\right)
=
\nabla_{x_t}\log \int p\left(y\mid x_0\right)p\left(x_0\mid x_t\right)\operatorname{d}\!x_0.
\label{eq:intractable_likelihood}
\end{equation}

\vspace{0.1em}
\begin{center}
    \includegraphics[width=0.95\linewidth,height=0.4\textheight,keepaspectratio]{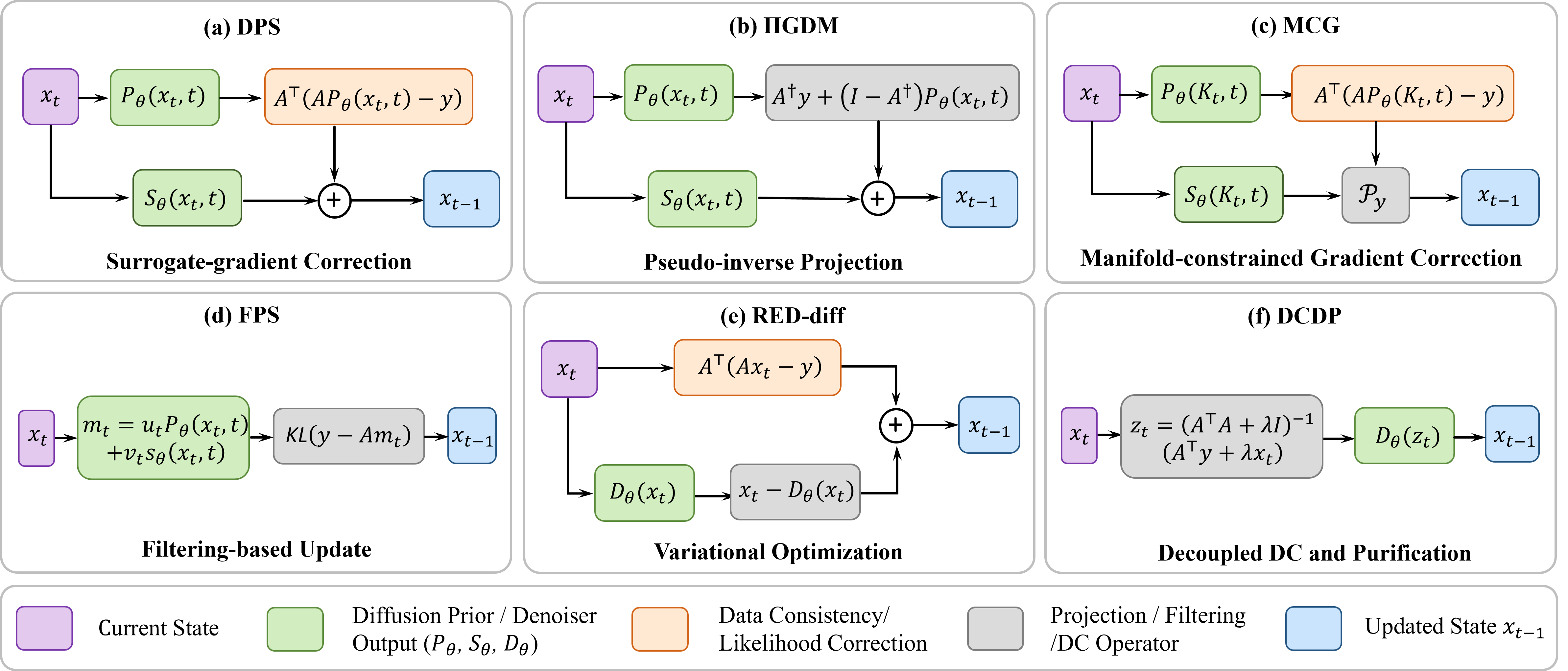}
    \captionof{figure}{Iterative pipelines of six representative diffusion-based methods for inverse problems.}
    \label{fig:diffusion_solver_pipelines}
\end{center}

A widely used strategy is to approximate the clean signal $x_0$ by the denoised prediction $P_\theta\left(x_t,t\right)$. Based on this plug-in approximation, Diffusion Posterior Sampling (DPS) \citep{Chung2023} evaluates data consistency in the denoised signal space. This leads to the update
\begin{equation}
x_{t-1}
=
x_t
-
\lambda A^\top
\left(
AP_\theta\left(x_t,t\right)-y
\right)
-
g^2\left(t\right)s_\theta\left(x_t,t\right),
\label{eq:dps_update}
\end{equation}
where $A^\top$ represents the transpose of the forward operator. As shown in Fig.~\ref{fig:diffusion_solver_pipelines}(a), DPS contains two parallel prior-related branches: the denoised prediction $P_\theta\left(x_t,t\right)$ is used to construct the surrogate measurement residual $A^\top\left(AP_\theta\left(x_t,t\right)-y\right)$, while the score $s_\theta\left(x_t,t\right)$ contributes to the reverse diffusion update. The resulting correction is therefore a surrogate-gradient correction, because the likelihood gradient is evaluated through the predicted clean signal rather than directly through the current state.

Building on a similar use of the denoised prediction, Pseudoinverse-Guided Diffusion Models ($\Pi$GDM) \citep{Song2023} further exploit the algebraic structure of the forward operator . Instead of back-projecting the measurement residual as in DPS, $\Pi$GDM decomposes the reconstruction into a measured component and a null-space component:
\begin{equation}
x_{t-1}
=
A^\dagger y
+
\left(I-A^\dagger A\right)P_\theta\left(x_t,t\right)
-
g^2\left(t\right)s_\theta\left(x_t,t\right),
\label{eq:pigdm_update}
\end{equation}
where $A^\dagger$ denotes the pseudo-inverse of the forward operator. As illustrated in Fig.~\ref{fig:diffusion_solver_pipelines}(b), the observed subspace is constrained by $A^\dagger y$, while the unconstrained null-space component is supplied by the denoised prediction $P_\theta\left(x_t,t\right)$. The score term $s_\theta\left(x_t,t\right)$ still participates in the reverse update, while the data-consistency mechanism is dominated by the pseudo-inverse projection. This makes $\Pi$GDM a projection-guided method whose denoising error mainly propagates through the null-space component.

In contrast to projection-free posterior guidance strategies, Manifold-Constrained Guidance (MCG) \citep{Chung2022} introduces a manifold-constrained measurement-gradient correction through the Tweedie denoised estimate. Specifically, MCG first maps the current noisy iterate $x_t$ to a clean estimate $P_\theta\left(x_t,t\right)$, and then evaluates the measurement residual in this denoised space while taking the gradient with respect to $x_t$:
\begin{equation}
x_{t-1}^{\prime}
=
x_t
-
\eta A^\top
\left(
AP_\theta\left(x_t,t\right)-y
\right)
-
g^2\left(t\right)s_\theta\left(x_t,t\right),
\quad
x_{t-1}
=
\mathcal{P}_y\left(x_{t-1}^{\prime}\right).
\label{eq:mcg_update}
\end{equation}
Here, $\mathcal{P}_y\left(\cdot\right)$ denotes a projection-based data-consistency operator associated with the measurement $y$. It enforces consistency with the observed measurements after the manifold-constrained gradient correction. As shown in Fig.~\ref{fig:diffusion_solver_pipelines}(c), the data-consistency branch of MCG is therefore driven by the denoised-space residual $AP_\theta\left(x_t,t\right)-y$. The key distinction from DPS is that MCG further applies an explicit projection-based data-consistency step after the manifold-constrained gradient correction. This design encourages the diffusion trajectory to remain close to the learned data manifold while enforcing measurement consistency through $\mathcal{P}_y\left(\cdot\right)$.

From another perspective, Filtering Posterior Sampling (FPS) \citep{Dou2024} formulates diffusion-based inverse solving as a filtering posterior sampling process. Instead of constructing a deterministic likelihood-gradient approximation, FPS updates the posterior by combining diffusion-driven transition with measurement likelihood information:
\begin{equation}
p_\theta\left(x_{t-1}\mid x_t,y\right)
\propto
p_\theta\left(x_{t-1}\mid x_t\right)
p\left(y\mid x_{t-1}\right).
\label{eq:fps_posterior}
\end{equation}
In practical implementations, this recursive posterior update can be expressed as a gain-weighted correction:
\begin{equation}
x_{t-1}
=
m_t
+
KL\left(y-Am_t\right),
\quad
m_t
=
u_tP_\theta\left(x_t,t\right)
+
v_ts_\theta\left(x_t,t\right),
\label{eq:fps_update}
\end{equation}
where the intermediate prediction $m_t$ combines the denoised estimate $P_\theta\left(x_t,t\right)$ and the score information $s_\theta\left(x_t,t\right)$, and the gain term controls how strongly the measurement residual contributes to the update. As illustrated in Fig.~\ref{fig:diffusion_solver_pipelines}(d), FPS first forms a diffusion-based prediction $m_t$, and then corrects it through a gain-weighted measurement update. Although this correction is Kalman-like in form, FPS should be understood as a diffusion posterior sampling method with filtering-based measurement incorporation, rather than as a conventional Kalman filter.

In contrast to posterior-sampling-based approaches, Regularization by Denoising Diffusion Process (RED-diff) \citep{Mardani2024} adopts a variational optimization perspective. Instead of using the denoised prediction only as a likelihood surrogate, RED-diff introduces an implicit regularization term defined by a diffusion denoiser:
\begin{equation}
\min_x
\left\|Ax-y\right\|^2
+
\lambda R\left(x\right),
\quad
\nabla R\left(x\right)
\approx
x-D_\theta\left(x\right).
\label{eq:reddiff_objective}
\end{equation}
This yields the iterative update:
\begin{equation}
x_{t-1}
=
x_t
-
\eta A^\top\left(Ax_t-y\right)
-
\eta\lambda
\left(
x_t-D_\theta\left(x_t\right)
\right),
\label{eq:reddiff_update}
\end{equation}
where $D_\theta\left(\cdot\right)$ denotes the diffusion denoiser. As shown in Fig.~\ref{fig:diffusion_solver_pipelines}(e), RED-diff contains two coupled descent branches: the data-fidelity gradient $A^\top\left(Ax_t-y\right)$ and the denoiser-induced regularization term $x_t-D_\theta\left(x_t\right)$. These two components are combined within the same update step. Therefore, unlike DPS or $\Pi$GDM, where the denoised prediction mainly serves as a surrogate or null-space component, RED-diff directly embeds the denoiser into the optimization direction.

Finally, decoupled Data Consistency via Diffusion Purification (DCDP) \citep{Li2024} explicitly decouples data consistency and diffusion prior refinement into two successive operators. Instead of combining measurement correction and diffusion prior guidance in a single update, DCDP first performs a closed-form data-consistency projection and then applies diffusion-based purification:
\begin{equation}
z_t
=
\left(A^\top A+\lambda I\right)^{-1}
\left(A^\top y+\lambda x_t\right),
\quad
x_{t-1}
=
D_\theta\left(z_t\right).
\label{eq:dcdp_update}
\end{equation}
As shown in Fig.~\ref{fig:diffusion_solver_pipelines}(f), the gray block corresponds to the data-consistency projection, while the green block $D_\theta\left(\cdot\right)$ performs diffusion purification on the projected variable. This alternating structure separates measurement enforcement from generative refinement, making DCDP different from RED-diff, where data fidelity and denoiser regularization are coupled in a joint descent step.

\vspace{0.6em}
\begin{center}
    \includegraphics[width=0.95\linewidth,height=0.45\textheight,keepaspectratio]{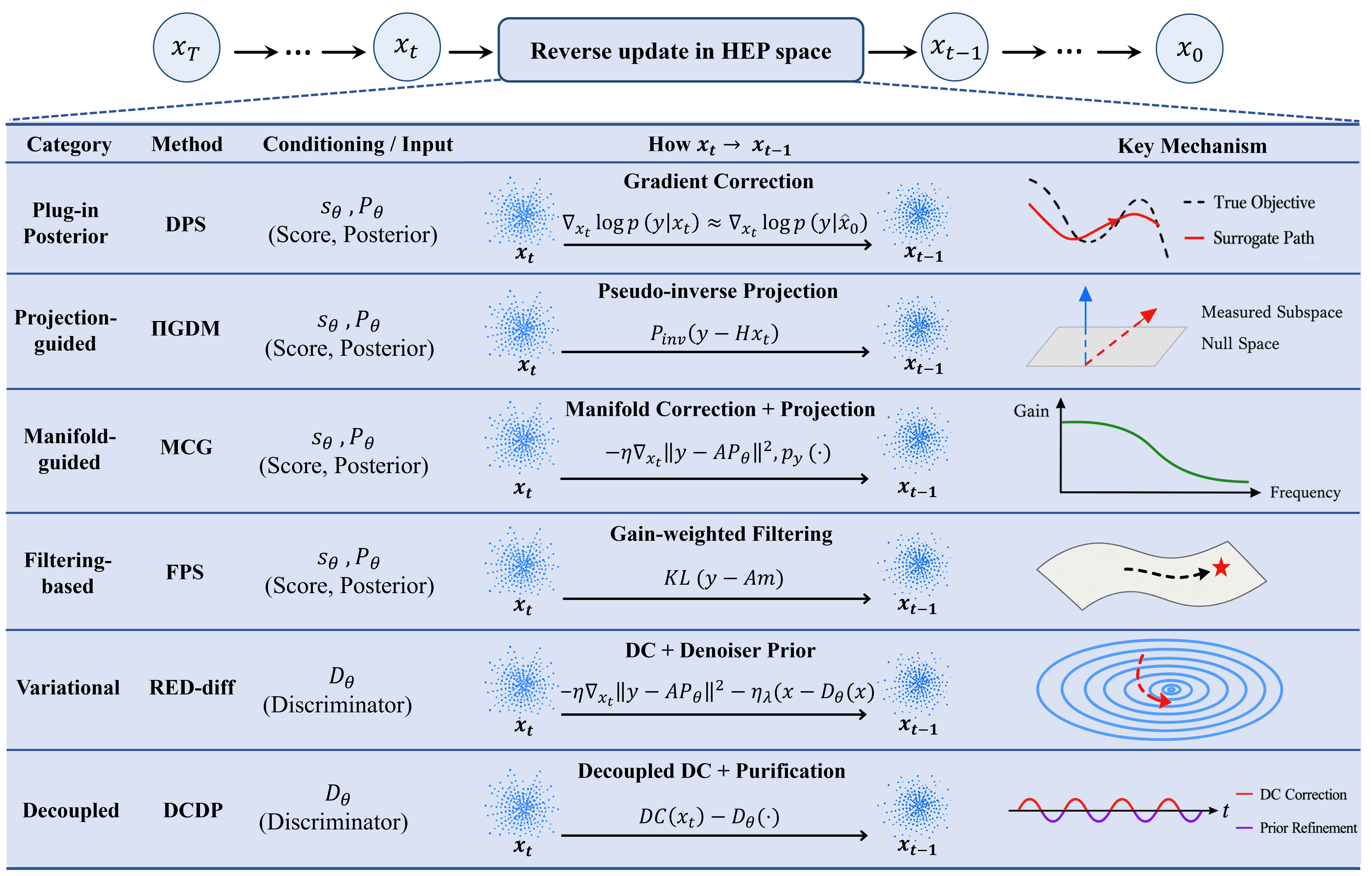}
    \captionof{figure}{Category-wise comparison of representative diffusion inverse solvers from the perspectives of conditioning input, $x_t\rightarrow x_{t-1}$ update formulation, and core optimization mechanism.}
    \label{fig:solver_category_comparison}
\end{center}
\vspace{0.4em}

The six solvers discussed above can be further compared from an operator-level perspective. As shown in Fig.~\ref{fig:solver_category_comparison}, all methods follow a common reverse diffusion backbone from $x_t$ to $x_{t-1}$, but they differ in their conditioning information, update mechanism, geometric interpretation, and dominant error or optimization signal. DPS and $\Pi$GDM both rely on $s_\theta$ and $P_\theta$, but DPS propagates errors through a surrogate gradient, whereas $\Pi$GDM confines part of the update to the measured and null-space components. FPS modulates the update through a gain-controlled filtering step, while MCG performs manifold-constrained measurement guidance. RED-diff directly couples denoiser-induced regularization with data fidelity, whereas DCDP separates data consistency and prior refinement into alternating operations.

\section{Methodology}\label{sec:methodology}

The diffusion-based solvers reviewed in Section~\ref{subsec:diffusion-posterior-sampling} were originally developed mainly for image-domain inverse problems. However, MRI reconstruction is naturally formulated in k-space domain, where the measured frequency coefficients are directly acquired. In this section, we instantiate these representative inference strategies directly in k-space domain without altering their underlying probabilistic mechanisms. Specifically, each method retains its original inference objective and posterior handling strategy, while the reconstruction variable is defined over complex-valued k-space signals. This unified formulation allows us to systematically investigate how different diffusion-based inverse solvers behave in noisy k-space reconstruction under a consistent experimental protocol.

\subsection{Formulation of Noisy K-space Reconstruction}\label{subsec:kspace-formulation}

MRI reconstruction can be expressed directly in k-space domain. Let $k_0$ denote the fully sampled k-space corresponding to the underlying image. Since the acquisition process fundamentally consists of selecting a subset of frequency coefficients, accelerated MRI can be modeled as applying a sampling mask to $k_0$ \citep{Block2007,Uecker2014}. Mathematically, the noisy undersampled k-space acquisition is formulated as
\begin{equation}
k = Mk_0 + n,
\label{eq:kspace_forward_model}
\end{equation}
where $k$ represents the acquired undersampled k-space data, $M$ denotes the sampling mask, and $n$ denotes measurement noise \citep{Fessler2010,Liang2015,Tezcan2022}. This formulation makes explicit that MRI reconstruction can be interpreted as a missing-data recovery problem in the k-space domain, where the goal is to infer the unobserved frequency coefficients while remaining consistent with the acquired measurements.

\vspace{0.4em}
\vspace{0.4em}
\begin{center} \begin{minipage}{0.85\linewidth} \centering \includegraphics[width=\linewidth]{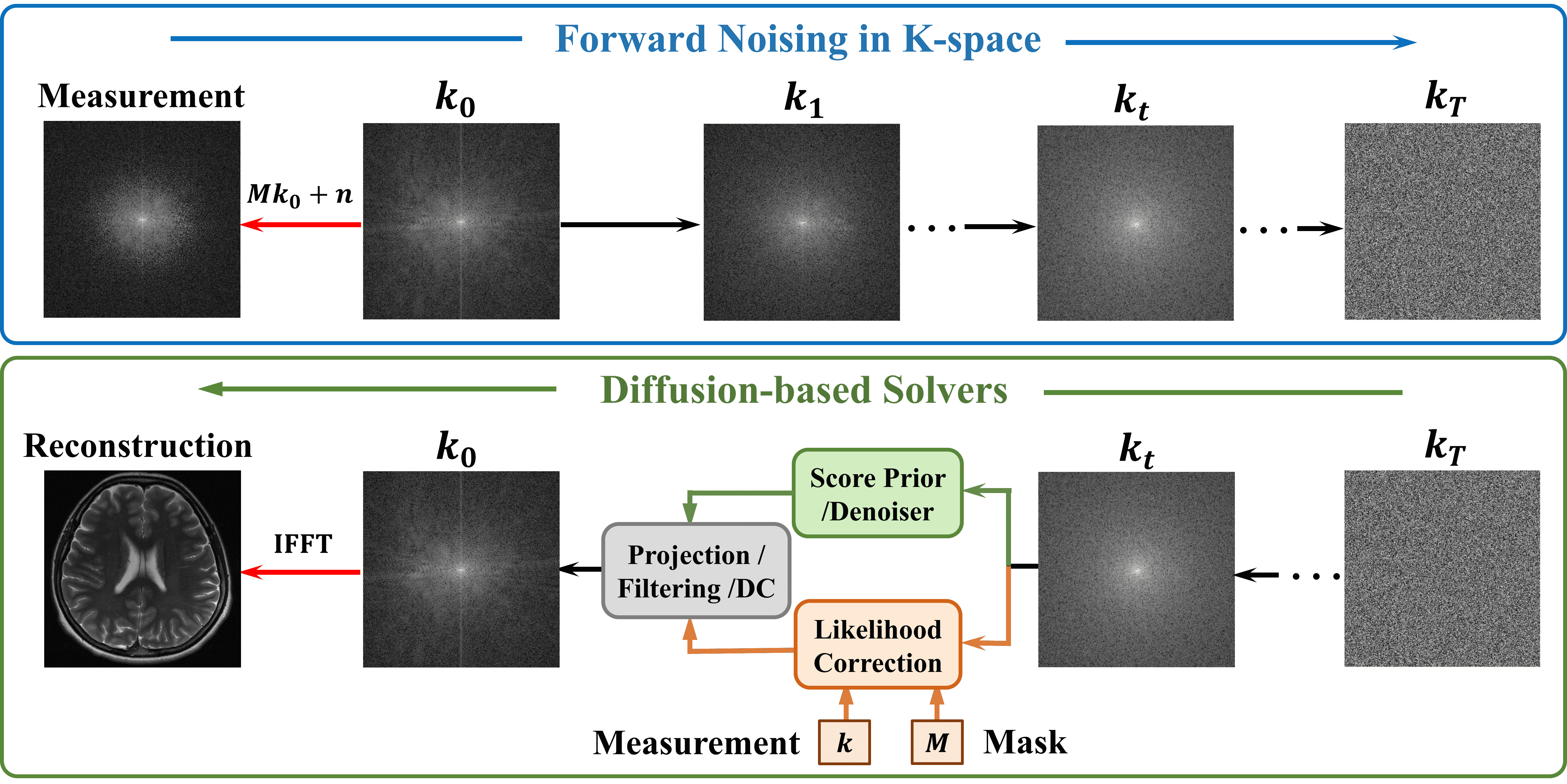} \captionof{figure}{Forward and reverse processes of the diffusion model in k-space domain.} \label{fig:kspace_diffusion_overview} \end{minipage} \end{center}
\vspace{0.7em}

An illustrative overview of the complete diffusion-based reconstruction process in the k-space domain is shown in Fig.~\ref{fig:kspace_diffusion_overview}. During the forward noising process, clean k-space data are progressively perturbed toward a fully noisy latent state. During the reverse denoising process, the learned diffusion prior is combined with measurement-related correction to recover the underlying fully sampled k-space signal \citep{Shen2024,Huang2023}. This formulation is consistent with the general posterior sampling framework introduced in Section~\ref{subsec:diffusion-posterior-sampling}, but differs in that the reconstruction variable is defined in the frequency domain rather than in the standard image domain.

Recent works have begun to exploit this perspective by introducing generative prior directly in k-space. For example, weighted k-space generative modeling applies a frequency-dependent reweighting strategy to k-space data before learning a diffusion prior \citep{Tu2023}. This strategy emphasizes different frequency regions according to their statistical characteristics, enabling the model to better capture the structure of MR signals, such as the concentration of energy in low-frequency components and the sparsity of high-frequency details. In such methods, reconstruction is typically performed under a noise-free or mildly corrupted assumption, where the acquired k-space samples are enforced exactly and the prior is mainly used to recover the missing entries.

In clinical MRI practice, however, undersampling and measurement noise often occur simultaneously \citep{Hyun2018,Qin2018}. Under this realistic setting, the acquired k-space measurements are jointly degraded by missing samples and noise corruption, making the reconstruction problem substantially more challenging. Compared with the noise-free case, this joint degradation further exacerbates the ill-posed nature of the inverse problem in the frequency domain: not only are large portions of k-space missing, but the remaining acquired measurements may also be unreliable \citep{Chung2023,Huang2024,Chen2024}. This coupling impairs rigid data-consistency constraints, as noisy measurements yield misleading guidance instead of valid reconstruction cues.

Consequently, direct k-space reconstruction under noisy and undersampled conditions becomes highly sensitive to measurement perturbations and inference errors. This motivates a more expressive and structured prior that improves the stability of diffusion-based reconstruction without modifying the fundamental solver formulation \citep{Haldar2014,Shin2014,Jin2016}. In the following section, we introduce a high-dimensional representation for k-space reconstruction, which lifts the original k-space data into enriched representation spaces and provides a general mechanism for improving diffusion-based inverse solvers.

\subsection{High-dimensional Representation for Noisy K-space Reconstruction}\label{subsec:high-dimensional-representations}

Recent advances in MRI reconstruction have increasingly moved beyond the original image-space formulation, instead exploring high-dimensional representation to better characterize the complex structure of k-space data. A common motivation behind these methods is that k-space is inherently entangled, where spatial structure, coil sensitivities, and frequency-dependent correlations are coupled within a compact representation, making direct modeling challenging. To alleviate this limitation, recent works introduce explicit high-dimensional embeddings to decouple and re-express k-space data in more structured feature spaces. Representative methods include WKGM \citep{Tu2023}, HKGM \citep{Peng2023}, CM-DM \citep{Guan2024}, and Sub-DM \citep{Guan2026}. Although these methods differ in implementation, they share a common idea: constructing a high-dimensional representation of k-space before applying generative modeling. In this context, high-dimensionality does not refer to increasing spatial resolution, but to lifting k-space data into richer feature domains.

\vspace{0.8em}
\begin{center} \begin{minipage}{\linewidth} \centering \includegraphics[width=0.8\linewidth]{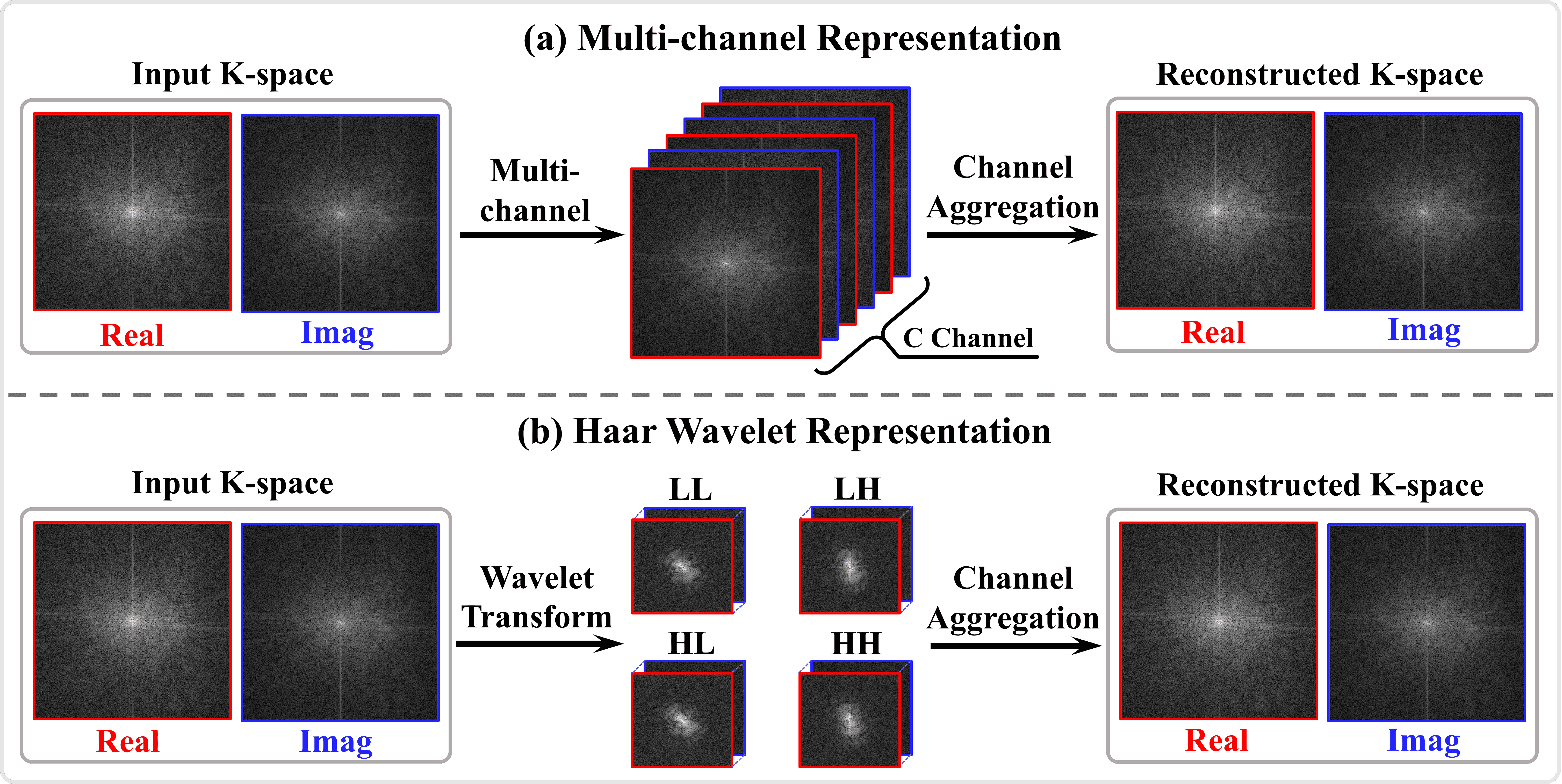} \captionof{figure}{Illustration of multi-channel and wavelet-transform HEP for complex-valued k-space. Real and imaginary components are lifted into high-dimensional representation and then aggregated back to original k-space domain.} \label{fig:high_dimensional_representations} \end{minipage} \end{center}
\vspace{0.7em}

As shown in Fig.~\ref{fig:high_dimensional_representations}(a), the proposed multi-channel extension explicitly constructs a rich high-dimensional representation by mapping the original complex-valued k-space signal to multi-channel components. Specifically, the input multi-channel k-space data $k$ is reorganized as
\begin{equation}
k \rightarrow K=\left\{k_1,k_2,\ldots,k_C\right\}\in\mathbb{C}^{C\times N},
\label{eq:mc_representation}
\end{equation}
where each channel corresponds to distinct coil acquisitions or separated real and imaginary signal components. This operation increases the representational capacity of the input by exposing inter-channel correlations that are not explicit in the original domain. After diffusion-based processing, the original signal is recovered through channel aggregation:
\begin{equation}
k=\frac{1}{C}\textstyle\sum_{i=1}^{C}K_i.
\label{eq:mc_aggregation}
\end{equation}
This bidirectional structure preserves physical consistency while enabling expressive modeling in the expanded space.

Fig.~\ref{fig:high_dimensional_representations}(b) also presents a wavelet transform embedding strategy, where k-space data is transformed into a multi-scale representation via discrete wavelet transform:
\begin{equation}
k \rightarrow K=W\left(k\right)=\left\{k^{\mathrm{LL}},k^{\mathrm{LH}},k^{\mathrm{HL}},k^{\mathrm{HH}}\right\}.
\label{eq:wavelet_representation}
\end{equation}
This decomposition separates low-frequency structural components from high-frequency directional details, enabling the diffusion model to operate on decorrelated subbands across different scales. After processing in the wavelet domain, the original k-space data is reconstructed through inverse transform:
\begin{equation}
k_{t-1}^{\prime}=W^{T}\left(K_{t-1}\right),
\label{eq:wavelet_inverse}
\end{equation}
which aggregates all subbands back into the full-frequency representation.

These transformations aim to reduce entanglement in original k-space domain and provide a more structured space for subsequent generative modeling \citep{Zhao2024,Peng2023}. From a modeling perspective, these approaches are typically developed under a diffusion-based framework, where a generative prior is learned in either image space or latent space, and reconstruction is achieved via projection back to satisfy measurement consistency \citep{Guan2024}. However, most of these formulations assume noise-free or mildly corrupted observations, and thus do not explicitly address the joint presence of undersampling and measurement noise. To address this limitation, we introduce a more general formulation that extends k-space data into a high-dimensional embedding space:
\begin{equation}
K=\mathrm{HEP}\left(k\right),
\label{eq:hep_mapping}
\end{equation}

where $k$ denotes the original k-space signal and $\mathrm{HEP}(\cdot)$ denotes the HEP operator. This framework unifies diffusion inverse solvers as instantiations on different representation domains: reconstruction in HEP-transformed k-space preserves each solver's native probabilistic inference while enhancing representation capacity. As detailed in Algorithm~\ref{alg:hep_reconstruction}, HEP serves as an external representation transform. Noisy undersampled k-space data are embedded into a high-dimensional space for diffusion posterior sampling with a target inverse solver, and the outputs are mapped back to raw k-space by inverse aggregation. The solvers' update rules remain unchanged, with score estimation, denoised prediction, and data-consistency correction performed in the HEP embedding.

\vspace{0.8em}
\begin{center}
\begin{minipage}{0.555\linewidth}
\refstepcounter{algorithm}
\label{alg:hep_reconstruction}
\noindent\rule{\linewidth}{1.2pt}
\vspace{1mm}
\noindent\textbf{Algorithm~\thealgorithm} HEP Framework for Noisy K-space Reconstruction
\vspace{1mm}
\hrule
\vspace{1mm}
\footnotesize
\begin{algorithmic}[1]
\Require $k_u^C,N,C,\mathrm{HEP},\mathrm{HEP}^{\dagger}$, Solver Type $ST$
\Statex \hspace{\algorithmicindent} \hspace*{0.5em} $ST\in\{\mathrm{DPS},\Pi\mathrm{GDM},\mathrm{FPS},\mathrm{MCG},\mathrm{RED\mbox{-}diff},\mathrm{DCDP}\}$

\State $K_u^C\gets \mathrm{HEP}\left(k_u^C\right)$
\State $K_N^C\sim\mathcal{N}\left(0,I\right)$
\For{$i\gets N-1$ \textbf{to} $0$}
    \For{$j=1$ \textbf{to} $C$}
        \State $\hat{\epsilon}_i^j\gets \epsilon_{\theta}\left(K_i^j\right)$
        \State $K_{i-1}^j\gets ST\left(K_u^C,K_i^j,\hat{\epsilon}_i^j\right)$
    \EndFor
\EndFor
\State $k_0^C\gets \mathrm{HEP}^{\dagger}\left(K_0^C\right)$
\State \Return $k_0^C$
\end{algorithmic}
%\vspace{1mm}
\noindent\rule{\linewidth}{1.2pt}
\end{minipage}
\end{center}
\vspace{0.9em}

Starting from the diffusion-based posterior sampling framework detailed in Section~\ref{subsec:diffusion-posterior-sampling}, we expand the original formulation originating from latent variable \(x_t\) into a high-dimensional k-space representation defined by \(K=\mathrm{HEP}\left(k\right)\). With this coordinate transformation, the inverse problem is restated as posterior sampling targeting the embedded variable \(K_t\), while retaining the identical probabilistic structure introduced in Section~\ref{subsec:diffusion-posterior-sampling}. The complete iterative posterior sampling workflow within HEP-transformed k-space is visualized in Fig.~\ref{fig:hep_iterative_framework}, which delivers a holistic overview of our proposed high-dimensional reconstruction framework.

Specifically, the posterior sampling in the high-dimensional space can be written in the same form as
\begin{equation}
\frac{\operatorname{d}\!K_t}{\operatorname{d}\!t}
=
f\left(K_t,t\right)
-
g^2\left(t\right)
\left[
s_{\theta}\left(K_t,t\right)
-
\nabla_{K_t}\log p\left(K_u\mid K_t\right)
\right],
\label{eq:hep_posterior_sampling}
\end{equation}
where $s_{\theta}\left(K_t,t\right)$ denotes the score function defined in the HEP space, and $\nabla_{K_t}\log p\left(K_u\mid K_t\right)$ represents the likelihood term under the embedded representation. Compared with Section~\ref{subsec:diffusion-posterior-sampling}, both the prior and likelihood are evaluated in a structured high-dimensional k-space, partially decoupling coil-wise, spatial, and frequency dependencies.

Within this framework, diffusion-based reconstruction methods can be interpreted as different strategies for approximating the likelihood term $\nabla_{K_t}\log p\left(K_u\mid K_t\right)$ and integrating it into the reverse dynamics. All methods are unified as diffusion posterior inference in HEP-transformed k-space, differing mainly in likelihood approximation and constraint enforcement.

\vspace{0.4em}
\begin{center} 
\begin{minipage}{\linewidth}
\centering \includegraphics[width=\linewidth]{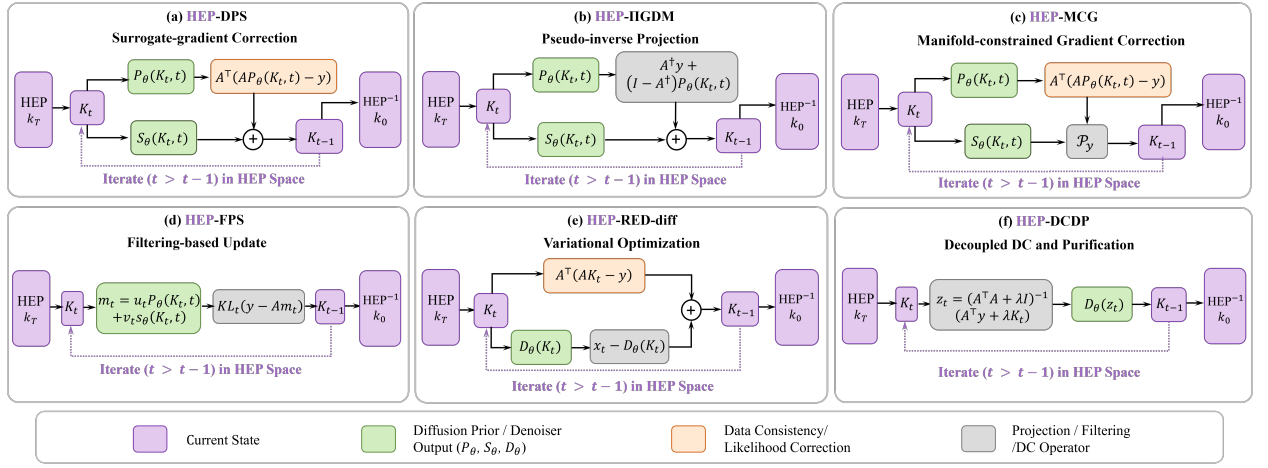} 
\captionof{figure}{Iterative reconstruction pipeline under the proposed high-dimensional k-space formulation $K=\mathrm{HEP}\left(k\right)$. Each subfigure illustrates iterative inference of a typical diffusion-based reconstruction solver in the embedded space.} \label{fig:hep_iterative_framework} 
\end{minipage} 
\end{center}
\vspace{0.6em}

As a representative instantiation, HEP-DPS is presented in Algorithm~\ref{alg:hep_dps}. DPS is chosen for its explicit residual back-projection guidance from the denoised prediction. Under HEP, this guidance is evaluated in structured high-dimensional k-space, while the original DPS sampling principle remains unchanged.

\vspace{0.2em}
\vspace{0.6em}
\begin{center}
\begin{minipage}{0.5\linewidth}
\refstepcounter{algorithm}
\label{alg:hep_dps}
{\hrule height 1.2pt width \linewidth}
\vspace{1mm}
\noindent\textbf{Algorithm~\thealgorithm} HEP-DPS for Noisy K-space Reconstruction
\vspace{1mm}
\hrule
\vspace{1mm}
\footnotesize
\begin{algorithmic}[1]
\Require $k_u^C,\mathrm{HEP},\mathrm{HEP}^{\dagger},M,N,C,\left\{\tilde{\sigma}_i\right\}_{i=1}^{N},\zeta$

\State $K_u^C\gets \mathrm{HEP}\left(k_u^C\right)$
\State $K_N^C\sim\mathcal{N}\left(0,I\right)$
\For{$i\gets N-1$ \textbf{to} $0$}
    \For{$j=1$ \textbf{to} $C$}
        \State $\hat{\epsilon}_i^j\gets \epsilon_{\theta}\left(K_i^j\right)$
        \State $\hat{K}_0^j\gets P_{\theta}\left(K_i^j,\hat{\epsilon}_i^j\right)$
        \State $K_{i-1}^j\gets
        \frac{\sqrt{\alpha_i}\left(1-\bar{\alpha}_{i-1}\right)}
        {1-\bar{\alpha}_i}K_i^j
        +
        \frac{\sqrt{\bar{\alpha}_{i-1}}\beta_i}
        {1-\bar{\alpha}_i}\hat{K}_0^j
        +
        \tilde{\sigma}_i$
        \State $K_{i-1}^j\gets K_{i-1}^j-\zeta_i\nabla_{K_i^j}
        \lVert y-M(\hat{K}_0^j)\rVert^2$
    \EndFor
\EndFor
\State $k_0^C\gets \mathrm{HEP}^{\dagger}\left(K_{i-1}^j\right)$
\State \Return $k_0^C$
\end{algorithmic}
\vspace{1mm}
{\hrule height 1.2pt width \linewidth}
\end{minipage}
\end{center}
\vspace{0.05em}

\subsection{Theoretical Mechanism: Error Reduction and Algorithmic Sensitivity}\label{subsec:theoretical-mechanism}

To elucidate why high-dimensional representation improve reconstruction, we analyze the problem from two coupled perspectives: (i) Error Reduction, which examines how HEP optimizes the statistical structure of score estimation; and (ii) Error Propagation, which explores how these improved prior are injected into various reconstruction solvers. This section analyzes the effectiveness of HEP in two parts. First, we show how HEP reduces the estimation error of the diffusion prior. Second, we examine how this reduced error impacts different reconstruction solvers, explaining why RED-diff benefits the most.

\subsubsection{\upshape Statistical Error Reduction via HEP}\label{subsubsec:statistical-error-reduction}

Recent studies have shown that diffusion models exhibit improved score estimation and distribution recovery when data lie in low-dimensional subspaces \citep{Chen2023}, and that generalization performance is strongly governed by the intrinsic dimension of embedding distributions rather than network size \citep{Yu2026}. Inspired by these findings, we derive the following error bound for high-dimensional embedding. We characterize the estimation error in the original k-space by its residual covariance matrix $\Sigma_k=\Sigma_{\mathrm{diag}}+\Sigma_{\mathrm{cross}}$, where $\Sigma_{\mathrm{diag}}$ represents independent dimensional variances and $\Sigma_{\mathrm{cross}}$ cross residual correction term.

\noindent\textbf{Theorem 1 (Error Bound of High-Dimensional Embedding).} Let $k\in\mathbb{C}^{N}$ be the original signal and $K=\mathrm{HEP}\left(k\right)\in\mathbb{R}^{C}\left(C\geq N\right)$ be a linear injective embedding. Given the score estimation risk $R=\mathbb{E}\left\|\delta\right\|^2$, the risks corresponding to the original space $R_k$ and the embedding space $R_K$ satisfy:
\begin{equation}
R_K \leq \frac{1}{\sigma_{\min}^{2}\left(\mathrm{HEP}\right)}R_k-\operatorname{Tr}\left(\Sigma_{\mathrm{cross}}\right)+\Delta_{\mathrm{HEP}},
\tag{23}
\end{equation}
where $\Delta_{\mathrm{HEP}}$ is the residual error introduced by the mapping.

\noindent\textbf{Proof:} Since HEP is a linear injective operator, there exists a bounded left inverse $\mathrm{HEP}^{\dagger}: k=\mathrm{HEP}^{\dagger}\left(K\right)$. The score or denoising error learned in the embedded space can be projected back to the original k-space through the inverse aggregation operator $\mathrm{HEP}^{\dagger}$:
\begin{equation}
s_K\left(K\right)=\left(\mathrm{HEP}^{\dagger}\right)^{*}s_k\left(k\right),
\tag{25}
\end{equation}
where $\left(\cdot\right)^{*}$ denotes the adjoint operator. $s_K$ denotes the ground truth score function in the embedding space, $\hat{s}_K$ represents the corresponding estimated score, and $\delta_K=\hat{s}_K-s_K$ denotes the score estimation error in the HEP space. We decompose the embedding space error $\delta_K$ as:
\begin{equation}
\delta_K=\left(\hat{s}_K-\left(\mathrm{HEP}^{\dagger}\right)^{*}\hat{s}_k\right)+\left(\mathrm{HEP}^{\dagger}\right)^{*}\delta_k.
\tag{26}
\end{equation}
Substituting Eq. (26) into the empirical risk $R_K=\mathbb{E}\lVert \delta_K \rVert^2$, we have

\begin{equation}
\begin{aligned}
R_K
=&\lVert \mathrm{HEP}^{\dagger *}\delta_k \rVert^2
+\lVert \hat{s}_K-\mathrm{HEP}^{\dagger *}\hat{s}_k \rVert^2&+2\langle \mathrm{HEP}^{\dagger *}\delta_k,\hat{s}_K-\mathrm{HEP}^{\dagger *}\hat{s}_k \rangle .
\end{aligned}
\tag{27}
\end{equation}

The first term is controlled by the spectral norm of $\mathrm{HEP}^{\dagger *}$. Since $\lVert \mathrm{HEP}^{\dagger *} \rVert^2=\lVert \mathrm{HEP}^{\dagger} \rVert^2=\frac{1}{\sigma_{\min}^{2}(\mathrm{HEP})}$, it can be described as:
\begin{equation}
\lVert \mathrm{HEP}^{\dagger *}\delta_k \rVert^2
\leq
\frac{1}{\sigma_{\min}^{2}(\mathrm{HEP})}
\lVert \delta_k \rVert^2
=
\frac{1}{\sigma_{\min}^{2}(\mathrm{HEP})}R_K.
\tag{28}
\end{equation}

Besides, the second term in Eq. (28) can be regarded as the HEP-induced mapping residual:
\begin{equation}
\Delta_{\mathrm{HEP}}
=
\lVert \hat{s}_K-\mathrm{HEP}^{\dagger *}\hat{s}_k \rVert^2.
\tag{29}
\end{equation}

The third term in Eq. (27) is the signed cross-residual interaction between the lifted original residual and the HEP-induced score correction. Then, we define
\begin{equation}
\operatorname{Tr}\left(\Sigma_{\mathrm{cross}}\right)
=
-2\mathbb{E}\left\langle\left(\mathrm{HEP}^{\dagger}\right)^{*}\delta_k,\hat{s}_K-\left(\mathrm{HEP}^{\dagger}\right)^{*}\hat{s}_k\right\rangle .
\tag{30}
\end{equation}
With this definition, a positive $\operatorname{Tr}\left(\Sigma_{\mathrm{cross}}\right)$ indicates that the HEP-induced correction reduces thelifted original residual in the empirical risk. Substituting Eqs. (28)--(30) into Eq. (27), we obtain
\begin{equation}
R_K
\leq
-\frac{1}{\sigma_{\min}^{2}\left(\mathrm{HEP}\right)}R_k
-
\operatorname{Tr}\left(\Sigma_{\mathrm{cross}}\right)
+
\Delta_{\mathrm{HEP}}.
\tag{31}
\end{equation}

\noindent\textbf{Corollary (Reconstruction Error Reduction).} The inverse diffusion iteration follows $x_{t-1}=f\left(x_t,s_0\left(x_t\right)\right)$, where $f\left(\cdot\right)$ denotes the inverse diffusion dynamics function that maps the current state and score estimate to the next state. The reconstruction error satisfies a recursive propagation rule: the error at step $t-1$ is determined by the error propagated from step $t$ and the score estimation error injected at current step.
\begin{equation}
\mathbb{E}\left\|\Delta x_{t-1}\right\|^2
\leq
c_t\mathbb{E}\left\|\hat{s}_{\theta}-s\right\|^2.
\tag{32}
\end{equation}
Recursive expansion yields the final error bound
\begin{equation}
\mathbb{E}\left\|\Delta x_0\right\|^2
\leq
\textstyle\left(\sum_{t=1}^{T}c_t\right)R.
\tag{33}
\end{equation}
The reconstruction errors in the original and embedding spaces are denoted as $\mathbb{E}\left\|\Delta k\right\|^2$ and $\mathbb{E}\left\|\Delta K\right\|^2$, respectively. Substituting Theorem 1 gives

\begin{equation}
\mathbb{E}\|\Delta K\|^2
\leq
\frac{1}{\sigma_{\min}^{2}(\mathrm{HEP})}
\mathbb{E}\|\Delta k\|^2
-
\operatorname{Tr}(\Sigma_{\mathrm{cross}})\textstyle\sum_{t=1}^{T}c_t
+
\Delta_{\mathrm{HEP}}\textstyle\sum_{t=1}^{T}c_t.
\tag{34}
\end{equation}

Define the error gain after embedding:
\begin{equation}
\Delta\varepsilon
=
\mathbb{E}\left\|\Delta K\right\|^2
-
\mathbb{E}\left\|\Delta k\right\|^2
=
\left(\frac{1}{\sigma_{\min}^{2}}-1\right)R_k\textstyle\sum c_t
+
\left(\Delta_{\mathrm{HEP}}-\operatorname{Tr}\left(\Sigma_{\mathrm{cross}}\right)\right)\textstyle\sum c_t.
\tag{35}
\end{equation}

The key term is therefore $\Delta_{\mathrm{HEP}}-\operatorname{Tr}\left(\Sigma_{\mathrm{cross}}\right)$. Because HEP is information-preserving, the HEP-space estimator can always degenerate to the transformed original estimator by Eq (25). For this degenerate case: $\hat{s}_K-\left(\mathrm{HEP}^{\dagger}\right)^{*}\hat{s}_k=0$ and thus $\Delta_{\mathrm{HEP}}-\operatorname{Tr}\left(\Sigma_{\mathrm{cross}}\right)=0$. Since the actual $\hat{s}_K$ is obtained by empirical risk minimization in the HEP space, its empirical risk should not exceed that of this degenerate solution. Therefore,
\begin{equation}
\Delta_{\mathrm{HEP}}-\operatorname{Tr}\left(\Sigma_{\mathrm{cross}}\right)\leq0.
\tag{36}
\end{equation}
If the HEP estimator learns a non-trivial correction beyond the degenerate solution, then
\begin{equation}
\Delta_{\mathrm{HEP}}-\operatorname{Tr}\left(\Sigma_{\mathrm{cross}}\right)<0.
\tag{37}
\end{equation}
Combining Eqs. (28) and (29), we obtain
\begin{equation}
R_K
\leq
\frac{1}{\sigma_{\min}^{2}\left(\mathrm{HEP}\right)}R_K.
\tag{38}
\end{equation}
For normalized channel-wise HEP or orthogonal transform-based HEP, $\sigma_{\min}\left(\mathrm{HEP}\right)\geq1$ holds, yielding $\frac{1}{\sigma_{\min}^{2}\left(\mathrm{HEP}\right)}-1\leq0$ intrinsically suppress reconstruction error. Therefore, HEP reduces the empirical score-estimation error and further propagates this reduction to the final reconstruction error.

\vspace{0.6em}
\begin{center} 
\begin{minipage}{\linewidth} \centering \includegraphics[width=\linewidth]{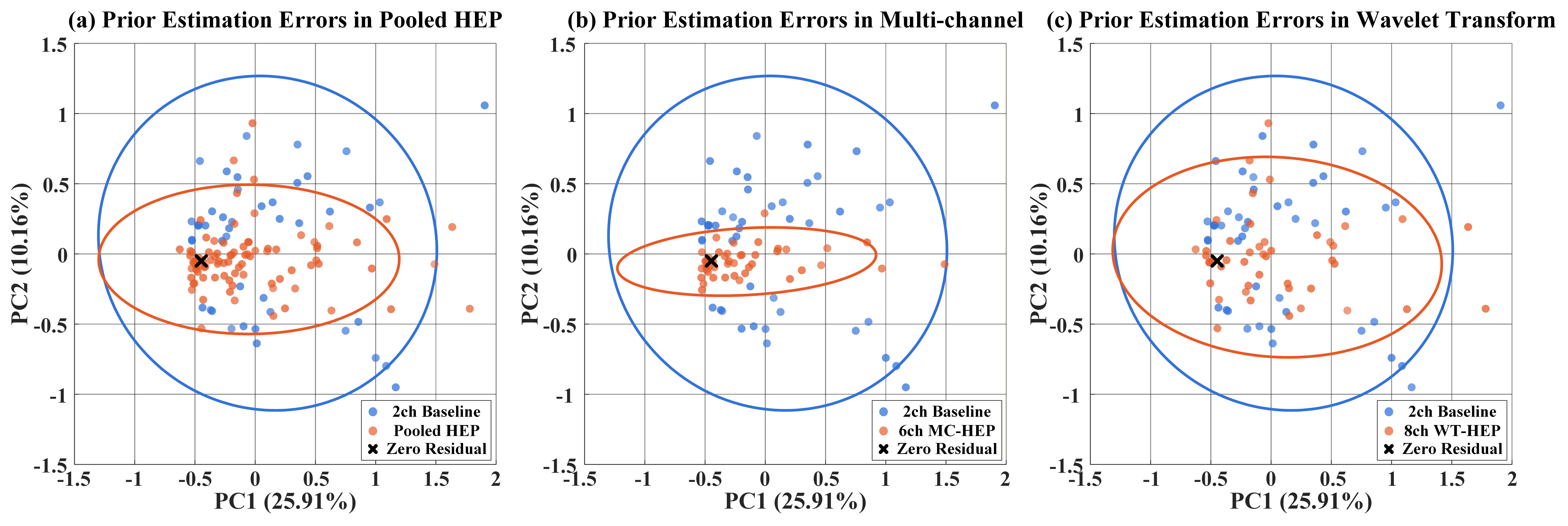} 
\captionof{figure}{PCA analysis of diffusion-prior estimation errors. Residuals are projected onto the first two principal components, and ellipses indicate 95\% confidence regions. HEP reduces residual dispersion, with multi-channel HEP showing the most compact distribution.} 
\label{fig:pca_prior_error} 
\end{minipage} 
\end{center}
\vspace{0.6em}

\noindent\textbf{Empirical Validation via PCA Analysis.} To investigate the validity of the derived error bounds, we performed a PCA-based analysis of the diffusion prior estimation residuals $\delta$. As shown in Fig.~\ref{fig:pca_prior_error}, the 2-channel baseline exhibits a widely dispersed residual distribution that deviates significantly from the zero-residual point, indicating strong cross-dimensional coupling in the original k-space representation. In contrast, HEP variants show a clear contraction of the 95\% confidence ellipses toward the origin, which empirically supports the correlation-suppression effect described in the theoretical analysis. Specifically, the multi-channel embedding yields the most compact and isotropic distribution, indicating that the gain from suppressing cross-dimensional coupling outweighs the residual mapping error $\Delta_{\mathrm{HEP}}$, leading to the smallest $\Delta\varepsilon$. The wavelet transform embedding also reduces the residual bias but exhibits a slightly larger spread, illustrating the trade-off between mapping residuals and correlation suppression. These observations show that HEP improves diffusion prior estimation in the embedded k-space representation.

\subsubsection{Algorithmic Error Sensitivity and Propagation}\label{subsubsec:algorithmic-error-sensitivity}

As demonstrated in Section~3.3.1, HEP consistently reduces the denoising risk $R_K$. However, whether this theoretical gain translates into improved final reconstruction accuracy depends on the propagation and modulation mechanisms of representation errors within specific algorithmic architectures.

\noindent\textbf{Theorem 2 (Algorithmic Sensitivity and Error Coupling).} Compared to solvers based on gradient guidance or null-space projection, variational solvers that tightly couple the regularization term with the data-fidelity term exhibit higher error sensitivity. Consequently, as the prior estimation error decreases, such coupled architectures achieve a more significant reduction in overall reconstruction error.

\noindent\textbf{Proof:} While Section~3.3.1 establishes that HEP consistently reduces the denoising risk $R_K$, the translation of this theoretical gain into final reconstruction fidelity is strictly governed by the error sensitivity of the specific algorithmic architecture. To formalize this, we define the denoising output of a solver at any timestep $t$ as:
\begin{equation}
P_{\theta}\left(K_t,t\right)=K^{*}+\delta_t,
\tag{39}
\end{equation}
where $K^{*}$ denotes the ground truth high-dimensional representation induced by $\mathrm{HEP}\left(\cdot\right)$, and $\delta_t$ is the representation error. Compared with the original image-space formulation, $\delta_t$ now reflects errors in the structured embedding space, including multi-channel and wavelet-induced components, and its propagation depends on how each method couples the diffusion prior with the measurement operator. For plug-in based posterior sampling methods such as DPS, the approximation for plug-in based posterior sampling methods such as DPS, the approximation $p\left(y\mid K_t\right)\approx P_{\theta}\left(K_t,t\right)$ leads to the gradient $-A^{\top}\left(AP_{\theta}\left(K_t,t\right)-y\right)$. Substituting the error decomposition yields the following result:
\begin{equation}
A^{\top}\left(AK^{*}-y\right)+A^{\top}A\delta_t.
\tag{40}
\end{equation}
Under approximate data consistency $AK^{*}\approx y$, the dominant term becomes $A^{\top}A\delta_t$, indicating that the reconstruction error depends linearly on the denoising error. Consequently, reducing $\delta_t$ through high-dimensional embedding directly improves performance in a first-order manner.

$\Pi$GDM, while structurally different, still relies on $P_{\theta}\left(K_t,t\right)$ through the null-space reconstruction
\begin{equation}
x_0=A^{\dagger}y+\left(I-A^{\dagger}A\right)P_{\theta}\left(K_t,t\right)-g^2\left(t\right)s_{\theta}\left(K_t,t\right).
\tag{41}
\end{equation}
Substituting the denoiser output, it gives
\begin{equation}
\left(I-A^{\dagger}\right)K^{*}+\left(I-A^{\dagger}\right)\delta_t,
\tag{42}
\end{equation}
It showing that the denoising error is confined to the null space of $A$. Since this component is inherently less constrained by the measurements, the reduction of $\delta_t$ only partially translates into reconstruction improvement, making the benefit of high-dimensional embedding weaker than in DPS.

In contrast, MCG implements a two-step iterative paradigm of gradient correction followed by projection-based data consistency, as formulated in Eq.~(10):
\begin{equation}
K_{t-1}^{\prime}=K_t-\eta A^{\top}\left(AP_{\theta}\left(K_t,t\right)-y\right)-g^2\left(t\right)s_{\theta}\left(K_t,t\right),\quad K_{t-1}=\mathcal{P}_y\left(K_{t-1}^{\prime}\right).
\tag{43}
\end{equation}
In this framework, the update is first refined by measurement gradients and time-varying constraints to reduce data-fitting deviations, and then calibrated by the projection operator $\mathcal{P}_y\left(\cdot\right)$ for strict data consistency. Unlike DPS and $\Pi$GDM, which mainly rely on surrogate-gradient guidance, MCG introduces post-gradient projection modulation to reduce deviations induced by denoising errors. This makes the benefit from error reduction less direct but enhances iterative robustness. Consequently, MCG, DPS, and $\Pi$GDM exhibit distinct strengths under noisy disturbances.

A further attenuation occurs in FPS, where the update takes the form
\begin{equation}
z_{t-1}=K_{t-1}+KL_t\left(y-AK_{t-1}\right),\quad K_t=u_tP_{\theta}\left(z_{t-1},t\right)+v_ts_{\theta}\left(z_{t-1},t\right).
\tag{44}
\end{equation}
Injecting the error into $m_t$ and expanding the correction term yields an error contribution proportional to $K_tAu_t$. The Kalman gain $K_t$, determined by the relative covariance structure, effectively rescales and often suppresses this term. As a result, the influence of $\delta_t$ is not only indirect but also modulated by the filtering mechanism, further diminishing the gains from improved denoising.

A similar decoupling is present in DCDP, where the reconstruction alternates between
\begin{equation}
z_{t+1}=\left(A^{\top}A+\lambda I\right)^{-1}\left(A^{\top}y+\lambda K_t\right),\quad K_{t+1}=D_{\theta}\left(z_{t+1}\right).
\tag{45}
\end{equation}
Letting the denoiser output, the data consistency step produces an error term filtered by $\left(A^{\top}A+\lambda I\right)^{-1}\lambda\delta_t$, which smooths and attenuates the denoising error before it is passed to the next iteration. The subsequent denoising step then reintroduces a new error $\delta_{t-1}$. This alternating structure effectively separates and weakens the influence of denoising accuracy, so variance reduction from high-dimensional embedding is largely dissipated across iterations.

Finally, RED-diff fundamentally embeds the denoiser into a variational objective, yielding the update
\begin{equation}
K_{t-1}=K_t-\eta A^{\top}\left(AK_t-y\right)-\eta\lambda\left(K_t-D_{\theta}\left(K_t\right)\right).
\tag{46}
\end{equation}
Substituting $D_{\theta}\left(K_t\right)=K^{*}+\delta_t$ gives
\begin{equation}
K_{t-1}=K_t-\eta A^{\top}\left(AK_t-y\right)-\eta\lambda\left[\left(K_t-K^{*}\right)-\delta_t\right]
\tag{47}
\end{equation}
The denoising error $\delta_t$ appears explicitly and additively in the update, and it is coupled with the deviation term $K_t-K^{*}$ within the same optimization step. Unlike previous methods where $\delta_t$ is projected, filtered, or temporally separated, RED-diff uses the denoiser as a direct surrogate for the prior gradient. Consequently, any reduction in $\delta_t$ simultaneously improves both the regularization force and the overall descent direction.

In summary, the extent to which an algorithmic architecture tolerates representation errors is determined by its underlying operator structure. By tightly coupling the diffusion prior with data constraints, RED-diff maximizes sensitivity to prior error reduction, thereby most effectively transforming the statistical gains of HEP into significant reconstruction performance enhancements.

\section{Experiments}\label{sec:experiments}

\subsection{Datasets}\label{subsec:datasets}

The proposed HEP unified reconstruction framework is evaluated on two MRI datasets, including an in-house dataset and a publicly available benchmark dataset for external validation.\textbf{In-house dataset:} The dataset was provided by the Shenzhen Institute of Advanced Technology (SIAT). Fully sampled multi-coil k-space data were acquired on a 3.0T Siemens Trio Tim MRI scanner using a 12-channel receiver coil. Each acquisition contains complex-valued k-space data with a matrix size of $256 \times 256$ and a field of view (FOV) of $220 \times 220~\mathrm{mm}^2$. The imaging protocol used a repetition time (TR) of 6100~ms and an echo time (TE) of 99~ms. A total of 500 slices were included in this dataset, among which 450 slices were used for training and 50 slices were used for testing. Random flipping and rotation were applied to the training set for data augmentation, resulting in 3600 training samples. For model training, the multi-coil data were converted into single-coil representations. \noindent\textbf{Public dataset:} To further assess generalization performance, experiments are conducted on the fastMRI \citep{Zbontar2018}. A subset of the brain dataset was used for evaluation. The fully sampled data were acquired with an encoded matrix size of $640 \times 320$ and reconstructed to images of size $320 \times 320$ with an FOV of $220 \times 220~\mathrm{mm}^2$. The dataset includes multi-coil acquisitions with 8 receiver channels and corresponds to a T1-weighted FLASH sequence (TR = 264~ms, TE = 2.88~ms, and flip angle = $70^\circ$). For consistency with the in-house dataset, all images were center-cropped to $256 \times 256$. Fully sampled k-space data were used as the ground truth for evaluation.

\subsection{Experimental Setup}\label{subsec:experimental-setup}

We consider MRI reconstruction in the k-space domain under an undersampled and noisy acquisition conditions, following the formulation introduced in Section 3. The diffusion prior is trained following the Improved Denoising Diffusion Probabilistic Models framework \citep{Nichol2021}. Specifically, the denoising network adopts a U-Net architecture with 128 channels and two residual blocks at each resolution level. The diffusion process is configured with 1000 timesteps, a linear noise schedule. Self-attention is applied at the $16 \times 16$ resolution using four attention heads, while scale-shift normalization and residual up/down-sampling are also enabled.

To handle complex-valued data, two representation strategies are investigated, including a 2-channel representation (real and imaginary parts), a 6-channel multi-channel embedding and an 8-channel wavelet  transform embedding. The model is optimized using Adam with a learning rate of $10^{-4}$, momentum coefficients $[0.9, 0.999]$ and weight decay of 0.1. The batch size is set to 4, and training is performed for 200000 iterations using a uniform timestep sampler, with exponential moving average (EMA) (decay rate 0.999) applied for stabilization. During inference, reconstruction is performed using standard reverse diffusion sampling with 1000 steps without timestep respacing.

All experiments are implemented in PyTorch and conducted on an NVIDIA RTX 5090 GPU (24 GB) under CUDA 12.8. Reconstruction performance is quantitatively evaluated using peak signal-to-noise ratio (PSNR), structural similarity index measure (SSIM) and normalized mean squared error (NMSE), computed in the image domain after inverse Fourier transformation. For all quantitative results, we report the average performance over multiple runs to reduce stochastic variability.

\subsection{Experimental Design}\label{subsec:experimental-design}

To comprehensively evaluate the proposed k-space reconstruction framework, we conduct a series of experiments under a unified setting to analyze the effects of inverse solvers, representation strategies, noise corruption, undersampling factors and classical prior integration. All experiments are performed on complex-valued measurements in the k-space domain under controlled degradation conditions.

We first evaluate the effectiveness of high-dimensional representations across six representative diffusion-based inverse solvers, including DPS, $\Pi$GDM, MCG, FPS, RED-diff and DCDP. All methods are tested under multiple undersampling ratios ($R=6, 10, 15$) with a fixed noise level of $\mathrm{SNR}=20~\mathrm{dB}$. Three representation strategies are considered, including 2-channel, multi-channel, and wavelet  transform embeddings. This experiment aims to verify the consistency of representation gains across different inverse solvers and varying undersampling conditions.

To further assess robustness under measurement degradation, we conduct experiments by varying both noise levels and undersampling factors. Specifically, noise robustness is evaluated under $\mathrm{SNR}=30~\mathrm{dB}$, $20~\mathrm{dB}$, and $10~\mathrm{dB}$ with a fixed undersampling ratio of $R=6$, while undersampling robustness is evaluated under $R=10, 20$, and $30$ with $\mathrm{SNR}=20~\mathrm{dB}$ Diffusion-based. These experiments examine the stability of different methods under progressively ill-posed reconstruction conditions. Finally, we explore the compatibility of high-dimensional representations with classical model-based prior by incorporating low-rank constraints into both multi-channel and wavelet transform embeddings. All experiments in this setting are conducted under $\mathrm{SNR}=20~\mathrm{dB}$ and $R=6, 10$, and $15$, enabling evaluation of whether traditional prior can further enhance representation-based reconstruction.

\section{Results}\label{sec:results}

We evaluate the proposed HEP framework from four complementary perspectives, designed to assess its effectiveness, robustness and compatibility with existing reconstruction strategies. First, we compare reconstruction performance with and without high-dimensional embedding across six representative diffusion-based inverse solvers. This experiment aims to determine whether HEP provides a consistent representation-level improvement, rather than gains tied to a specific solver formulation. Second, we evaluate robustness to measurement noise by conducting experiments under varying noise levels on the fastMRI dataset. This setting reflects more realistic acquisition conditions and examines the stability of the proposed framework under degraded observations. Third, to further assess performance under more challenging acquisition scenarios, we select RED-diff as a representative solver and evaluate it under higher undersampling factors and different sampling patterns. Finally, we investigate the compatibility of HEP with a classical low-rank structural prior to assess whether the proposed high-dimensional representation can be combined with conventional MRI regularization strategies. Quantitative evaluation is conducted using PSNR, SSIM, and NMSE, while qualitative comparisons are provided through visual inspection, including local magnified views and corresponding residual maps.

\subsection{HEP across Inverse Solvers}\label{subsec:hep-across-inverse-solvers}

We first evaluate whether the proposed HEP framework provides consistent improvements across different diffusion-based inverse solvers. Experiments are conducted on the SIAT in-house dataset under a fixed noise level of $\mathrm{SNR}=20~\mathrm{dB}$, with three acceleration factors ($\times 6$, $\times 10$, and $\times 15$). Six representative solvers are considered, including DPS, $\Pi$GDM, FPS, MCG, RED-diff and DCDP. For each solver, we compare the naive k-space representation without high-dimensional embedding against two HEP instantiations: multi-channel embedding and wavelet-domain embedding. Quantitative results are summarized in Table~\ref{tab:hep_solver_siat}.

Across all methods, reconstruction performance degrades as the acceleration factor increases, reflecting the increased ill-posedness induced by more aggressive k-space undersampling. Nevertheless, both HEP variants consistently improve reconstruction quality over the standard representation in most cases. This trend suggests that the benefit of HEP arises from representation-level enhancement, rather than from modifications to a specific inference mechanism. From a modeling perspective, these improvements are consistent with the analysis in Section~3.3.1, where HEP reduces diffusion prior estimation error by enhancing the statistical structure of the underlying representation. In particular, the multi-channel embedding often yields larger quantitative gains, indicating that expanding the channel dimension facilitates the modeling of structured correlations in complex-valued k-space data.

\vspace{0.4em}

\begin{center}
\vspace{0.4em}
\begin{minipage}{\linewidth}
\centering
\captionsetup{
    type=table,
    labelsep=newline,
    labelfont=bf
}
\captionof{table}{Reconstruction performance of the proposed HEP framework with different embedding strategies on the SIAT in-house brain dataset (fixed noise level SNR = 20 dB). Metrics are presented as PSNR (dB) / SSIM / NMSE ($10^{-2}$).}

\vspace{-0.5em}
\label{tab:hep_solver_siat}
\renewcommand{\arraystretch}{1.50}
{
\setlength{\doublerulesep}{1pt}
\resizebox{\linewidth}{!}{
\setlength{\tabcolsep}{5pt}
\begin{tabular}{>{\centering\arraybackslash}p{1.5cm}ccccccc}
\noalign{\hrule height 1.2pt}
\textbf{Pattern} & \textbf{AF} & \textbf{DPS} & \textbf{$\Pi$GDM} & \textbf{MCG} & \textbf{FPS} & \textbf{RED-Diff} & \textbf{DCDP} \\
\hline
\multirow[c]{3}{*}{Naive}
& $R=6$  & 31.59/0.875/0.736 & \textbf{32.18}/\textbf{0.895}/0.632 & 31.26/0.871/0.756 & 32.11/0.879/\textbf{0.627} & 32.12/0.865/0.704 & 30.37/0.873/1.618 \\
& $R=10$ & 30.45/0.845/1.673 & 31.10/0.863/1.578 & 30.60/0.844/1.693 & 31.32/0.864/1.589 & \textbf{31.39}/\textbf{0.865}/\textbf{1.483} & 29.16/0.845/1.664 \\
& $R=15$ & 30.13/0.834/1.820 & 30.75/0.854/1.612 & 29.45/0.835/1.984 & 30.76/0.857/1.627 & \textbf{30.78}/\textbf{0.858}/\textbf{1.601} & 27.78/0.815/3.364 \\
\noalign{\hrule height 1.2pt}
\textbf{Pattern} & \textbf{AF} & \textbf{{\color{heppurple}HEP-}DPS} & \textbf{{\color{heppurple}HEP-}$\Pi$GDM} & \textbf{{\color{heppurple}HEP-}MCG} & \textbf{{\color{heppurple}HEP-}FPS} & \textbf{{\color{heppurple}HEP-}RED-Diff} & \textbf{{\color{heppurple}HEP-}DCDP} \\
\noalign{\hrule height 1.2pt}
\multirow[c]{3}{*}{\makecell[c]{Multi-\\Channel}}
& $R=6$  & 32.95/0.881/1.065 & 33.86/0.907/0.586 & 32.89/0.897/0.599 & 33.61/0.901/0.495 & \textbf{34.26}/\textbf{0.916}/\textbf{0.420} & 32.73/0.878/1.013 \\
& $R=10$ & 31.61/0.858/1.451 & 32.31/0.888/0.703 & 31.53/0.868/1.579 & 32.82/0.891/0.670 & \textbf{32.94}/\textbf{0.903}/\textbf{0.533} & 30.56/0.848/1.664 \\
& $R=15$ & 30.81/0.857/1.727 & 31.10/0.863/\textbf{1.578} & 30.15/0.853/1.798 & 31.88/0.878/1.632 & \textbf{32.08}/\textbf{0.889}/1.601 & 29.89/0.833/2.849 \\
\hline
\multirow[c]{3}{*}{\makecell[c]{Wavelet\\Transform}}
& $R=6$  & 32.75/0.878/1.118 & 33.68/0.897/0.598 & 32.79/0.889/0.612 & 33.58/0.898/0.512 & \textbf{34.01}/\textbf{0.910}/\textbf{0.413} & 31.01/0.859/0.845 \\
& $R=10$ & 31.58/0.855/1.496 & 32.08/0.883/0.721 & 31.35/0.866/1.601 & 32.46/0.889/0.701 & \textbf{32.65}/\textbf{0.891}/\textbf{0.541} & 30.26/0.843/1.767 \\
& $R=15$ & 30.77/0.857/1.774 & 31.01/0.868/\textbf{1.609} & 30.02/0.848/1.809 & 31.76/0.854/1.618 & \textbf{32.01}/\textbf{0.886}/1.631 & 29.01/0.832/2.912 \\
\noalign{\hrule height 1.2pt}
\end{tabular}
}
}
\end{minipage}
%\vspace{0.1em}
\end{center}

\vspace{0.15em}
\begin{center}
    \includegraphics[width=0.9\linewidth]{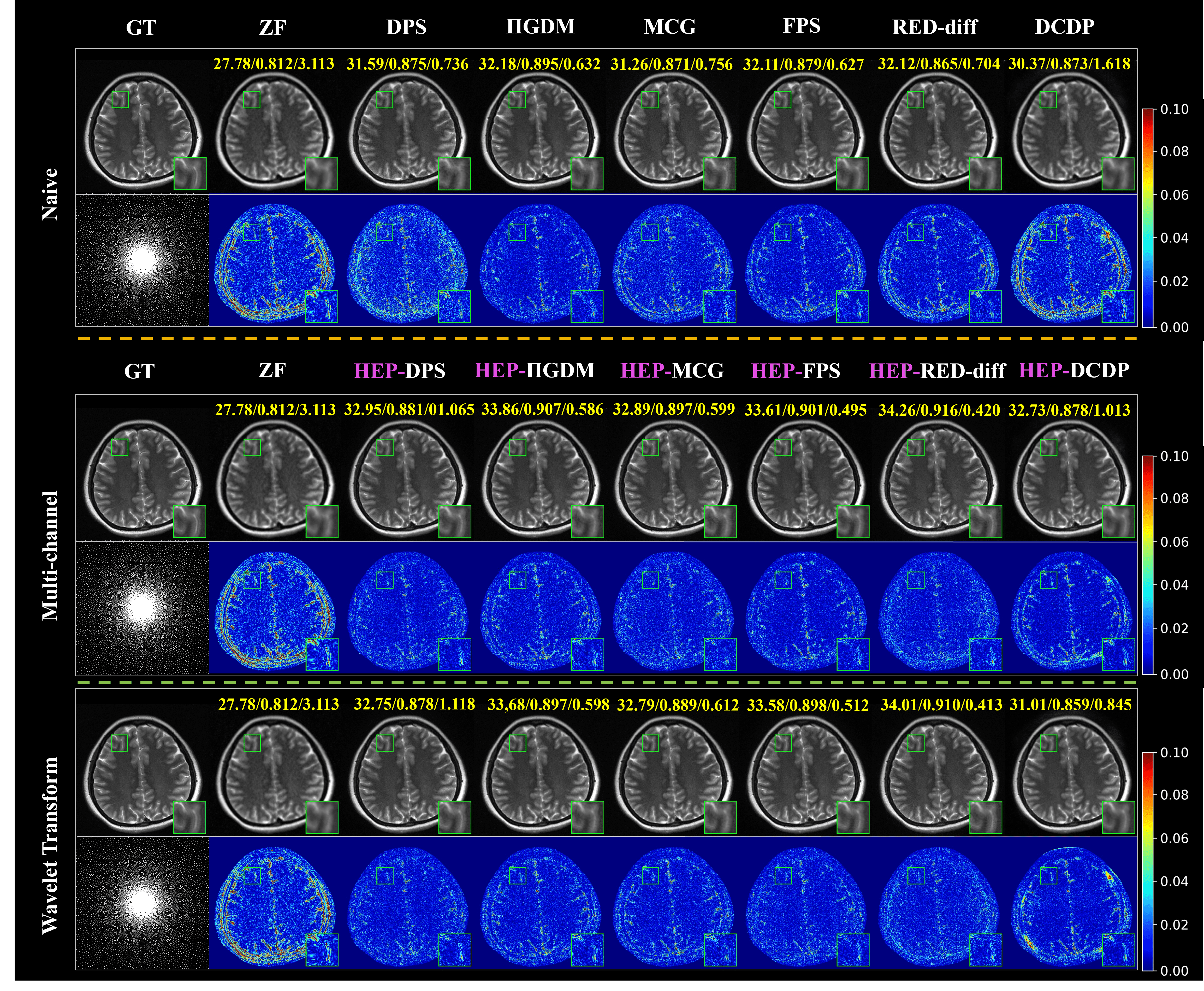}
    \captionof{figure}{Visual comparison of SIAT reconstructions under SNR = 20 dB and \(\times6\) acceleration. Naive, multi-channel HEP, and wavelet-transform HEP results are shown with zoomed regions and residual maps.}
    \label{fig:siat_qualitative_comparison}
\end{center}
\vspace{0.5em}

Among the evaluated solvers, RED-diff exhibits the most consistent performance improvement with HEP. This observation aligns with the error-propagation analysis in Section~3.3.2, where RED-diff tightly couples denoiser-based regularization with data-fidelity optimization, improvements in prior estimation can be more directly translated into reconstruction quality. In contrast, solvers based on filtering, projection, or alternating updates may partially attenuate such gains due to intermediate correction steps. Qualitative results in Fig.~\ref{fig:siat_qualitative_comparison} further corroborate these findings. Compared with reconstructions without high-dimensional embedding, HEP-based methods produce fewer residual artifacts and better preserve fine anatomical structures in the magnified regions. Overall, these results demonstrate that HEP serves as a general and effective representation enhancement for diffusion-based k-space MRI reconstruction, consistently benefiting a diverse set of inverse solvers.

\subsection{Robustness under Noise Variation}\label{subsec:robustness-noise-variation}

We further evaluate the robustness of the proposed HEP framework under varying measurement noise levels on the fastMRI dataset. In practical MRI acquisition, measurement noise is inevitable and perturbs the observed k-space coefficients, particularly in the presence of undersampling. Under this condition, the reconstruction problem becomes more challenging, as the algorithm must simultaneously recover missing frequency components and mitigate the influence of corrupted measurements. All experiments are conducted under a fixed acceleration factor of $\times 6$, with three noise levels corresponding to SNR values of $10~\mathrm{dB}$, $20~\mathrm{dB}$ and $30~\mathrm{dB}$. The same six diffusion-based inverse solvers are evaluated under both multi-channel and wavelet-domain HEP configurations, ensuring a consistent comparison across methods. Quantitative results are summarized in Fig.~\ref{fig:noise_variation_violin} using violin plots for PSNR, SSIM and NMSE, allowing both average performance and distributional stability to be compared under Poisson and radial sampling patterns.As shown in Fig.~\ref{fig:noise_variation_violin}, reconstruction performance decreases progressively as the noise level increases. Lower SNR values lead to reduced PSNR and SSIM, along with increased NMSE, reflecting the adverse impact of measurement noise on under-sampled k-space reconstruction. In this regime, noise not only corrupts the acquired frequency samples but also weakens the reliability of data-consistency constraints, thereby exacerbating the ill-posedness of the inverse problem.

\vspace{0.8em}
\begin{center}
\begin{minipage}{\linewidth}
    \centering
    \includegraphics[width=0.95\linewidth]{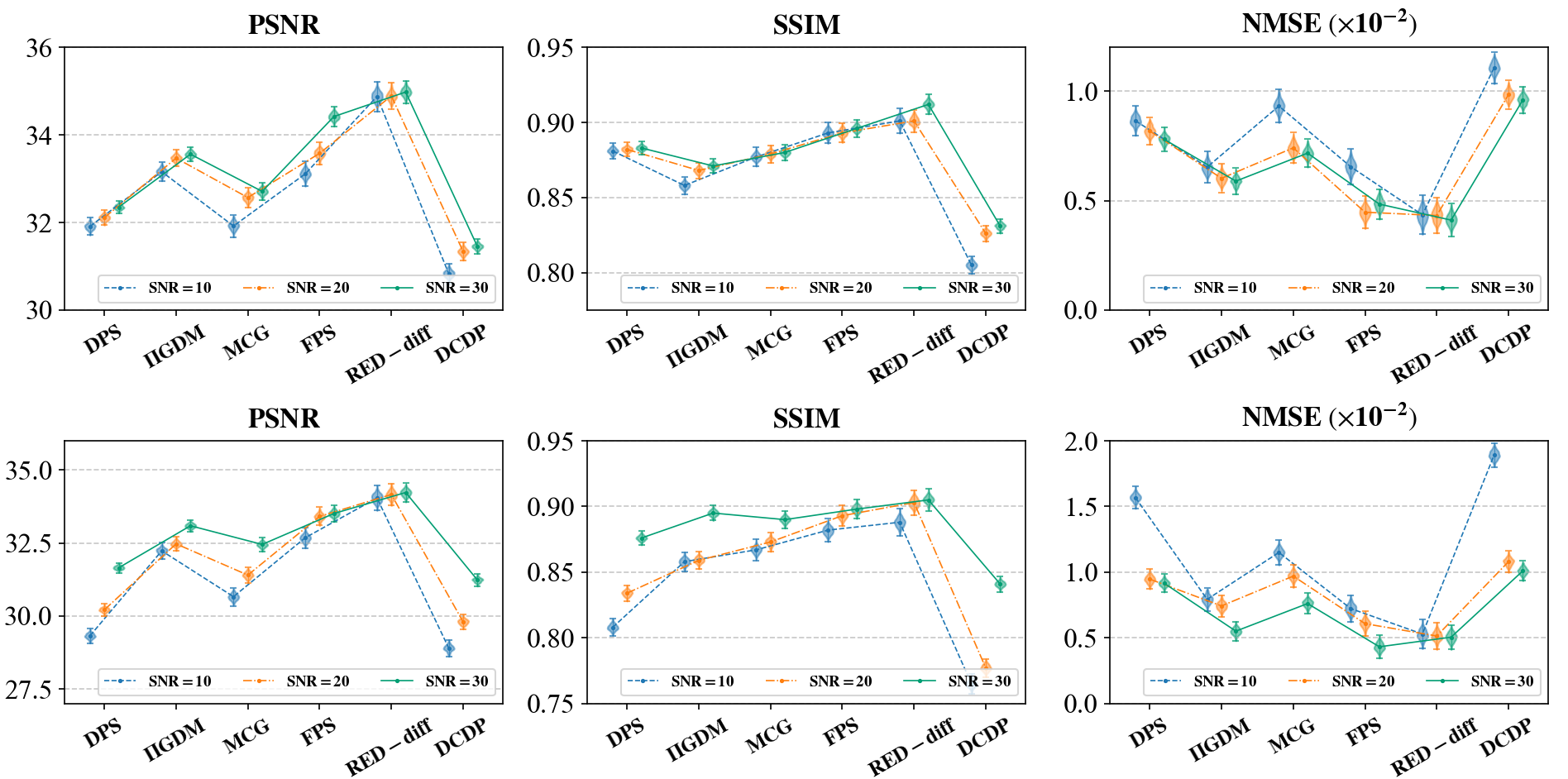}
    \captionsetup{
        type=figure,
        name=Fig.,
        labelfont=bf,
        labelsep=myperiod
    }
    \captionof{figure}{Statistical quantitative assessment of fastMRI reconstructions with $\times6$ acceleration under multiple noise conditions. Violin plots visualize PSNR, SSIM and NMSE distributions of competing solvers for Poisson and radial sampling.}
    \label{fig:noise_variation_violin}
\end{minipage}
\end{center}
\vspace{0.6em}

Despite this degradation, HEP-based methods exhibit more stable performance distributions across different noise levels, as evidenced by the reduced spread in the violin plots. This behavior indicates improved robustness to noise-corrupted k-space observations. From a modeling perspective, this advantage can be attributed to the structured high-dimensional embedding, which enables the diffusion prior to better capture the underlying signal distribution under degraded conditions, thereby mitigating the impact of noisy measurements. Among the evaluated solvers, RED-diff demonstrates consistently strong and stable performance across noise levels, supporting its use as a representative solver in subsequent experiments under more challenging undersampling settings. Qualitative results in Fig.~\ref{fig:noise_variation_qualitative} further corroborate these findings. Under noisy acquisition conditions, HEP-based reconstructions suppress residual artifacts and better preserve fine anatomical structures. Both multi-channel and wavelet-domain embeddings improve visual fidelity compared with reconstructions obtained without high-dimensional embedding.

\vspace{0.3em}
%\vspace{-0.2em}
\textbf{\begin{center}
\begin{minipage}{\linewidth}
    \centering
    \includegraphics[width=0.9\linewidth]{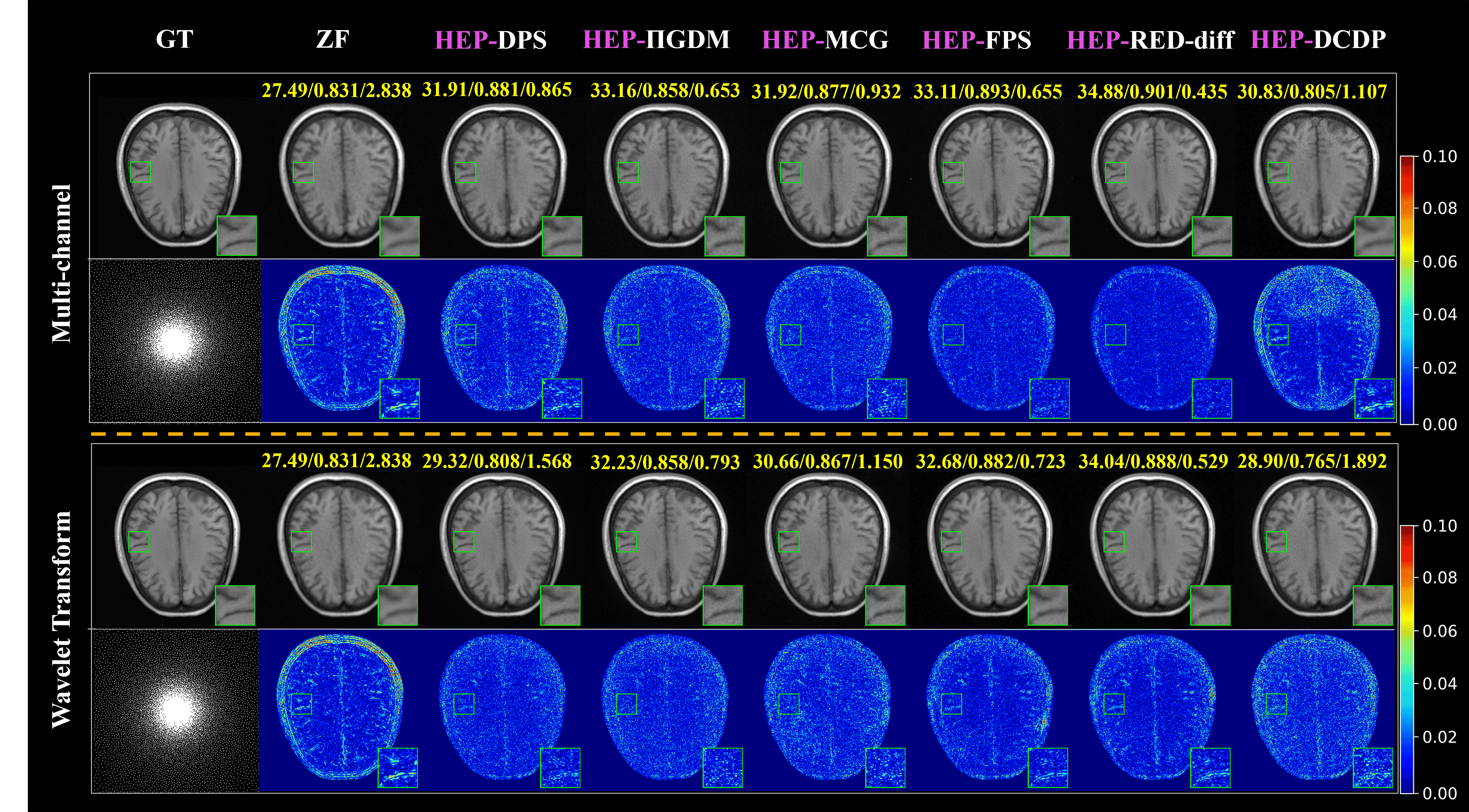}
    \captionsetup{
        type=figure,
        name=Fig.,
        labelfont=bf,
        labelsep=myperiod
    }
    \captionof{figure}{Qualitative comparison of HEP on fastMRI under SNR = 10 dB and $\times6$ acceleration. Multi-channel HEP and wavelet transform HEP results are shown with reconstructions, zoomed regions, and residual maps.}
    \label{fig:noise_variation_qualitative}
\end{minipage}
\end{center}}
\vspace{0.6em}
\vspace{0.2em}

\subsection{Effect of undersampling Factor}\label{subsec:effect-undersampling-factor}

Based on the results in Sections~5.1 and~5.2, RED-diff is selected as a representative solver for further evaluation under more challenging undersampling conditions. This selection is supported by its consistently strong empirical performance, as well as by the theoretical analysis in Section~3.3.2, which suggests that improvements in prior estimation can be more directly translated into reconstruction quality through its denoiser-induced regularization mechanism. To further examine the robustness of HEP-enhanced RED-diff under more aggressive acquisition settings, experiments are conducted on the fastMRI dataset with a fixed noise level of $\mathrm{SNR}=20~\mathrm{dB}$. The acceleration factors are increased to $\times 10$, $\times 20$ and $\times 30$, thereby inducing progressively more severe information loss in k-space. Two HEP-enhanced variants are evaluated, namely MC-RED-diff and WT-RED-diff, corresponding to multi-channel and wavelet transform embeddings, respectively. Compared with the previous experimental settings, this configuration presents a substantially more ill-posed reconstruction problem, as a larger proportion of k-space measurements is missing. Quantitative results are reported in Table~\ref{tab:undersampling_red_diff}.

\vspace{0.6em}

\begin{center}
\vspace{-0.5em}
\captionsetup{
    type=table,
    labelsep=newline,
    labelfont=bf
}
\captionof{table}{Numerical performance benchmark of MC-RED-diff and WT-RED-diff on fastMRI data at a fixed SNR of 20 dB with acceleration factors \(\times10\), \(\times20\) and \(\times30\). Reconstruction metrics include PSNR (dB), SSIM and NMSE ($10^{-2}$).}

\vspace{0.4em}
\label{tab:undersampling_red_diff}
\footnotesize
\setlength{\tabcolsep}{3pt}
\renewcommand{\arraystretch}{1.50}
\setlength{\doublerulesep}{1pt}

\begin{tabular}{>{\centering\arraybackslash}p{1cm}ccccccc}
\noalign{\hrule height 1.2pt}
\multirow[c]{2}{0.82cm}{\textbf{Pattern}} & \multirow[c]{2}{*}{\textbf{AF}} & \multicolumn{2}{c}{\textbf{{\color{heppurple}HEP-}RED-Diff}} \\
\noalign{\global\arrayrulewidth=0.4pt}\cline{3-4}\noalign{\global\arrayrulewidth=0.4pt}
% 修复这一行：\vline 放在 & 之间，分隔3、4列

& & \textbf{Multi-channel} & \textbf{Wavelet Transform} \\
% & & Multi-channel & Wavelet Transform \\

\hline
\multirow[c]{3}{*}{Poisson}
& $R=10$ & \textbf{31.86}/\textbf{0.873}/\textbf{0.894} & 31.70/0.849/1.073 \\
& $R=20$ & \textbf{30.01}/\textbf{0.831}/\textbf{1.370} & 28.34/0.803/2.009 \\
& $R=30$ & \textbf{25.13}/\textbf{0.734}/\textbf{4.820} & 24.75/0.724/5.612 \\
\hline
\multirow[c]{3}{*}{Radial}
& $R=10$ & \textbf{31.70}/\textbf{0.835}/\textbf{0.928} & 30.55/0.805/1.210 \\
& $R=20$ & \textbf{30.02}/\textbf{0.812}/\textbf{1.367} & 29.23/0.803/2.058 \\
& $R=30$ & \textbf{25.01}/\textbf{0.727}/\textbf{5.227} & 24.10/0.703/6.578 \\
\noalign{\hrule height 1.2pt}
\end{tabular}
\end{center}
\vspace{0.6em}
\vspace{0.3em}

As shown in Table~\ref{tab:undersampling_red_diff}, reconstruction performance decreases as the acceleration factor increases, reflecting the stronger ill-posedness introduced by more aggressive k-space undersampling. Nevertheless, both MC-RED-diff and WT-RED-diff maintain relatively stable performance trends at high acceleration factors, indicating that the proposed HEP framework remains effective beyond moderate undersampling regimes. A consistent performance gap is observed between the two variants, with MC-RED-diff achieving higher PSNR and SSIM and lower NMSE than WT-RED-diff across most settings. This result suggests that multi-channel embedding is more effective in preserving structural information in k-space under severe undersampling conditions. One possible explanation is that the multi-channel representation provides a richer embedding of channel-wise correlations while retaining a more direct mapping to the original k-space domain, thereby facilitating more accurate reconstruction. Representative qualitative results in Fig.~10 further show that high-dimensional RED-diff suppresses undersampling artifacts and preserves anatomical structures under both Poisson and radial sampling patterns. These results indicate that HEP-enhanced RED-diff remains robust under different sampling trajectories and more aggressive acceleration factors.

\vspace{0.4em}
\vspace{0.4em}
\begin{center}
    \begin{minipage}{\linewidth}
        \centering
        \includegraphics[width=\linewidth]{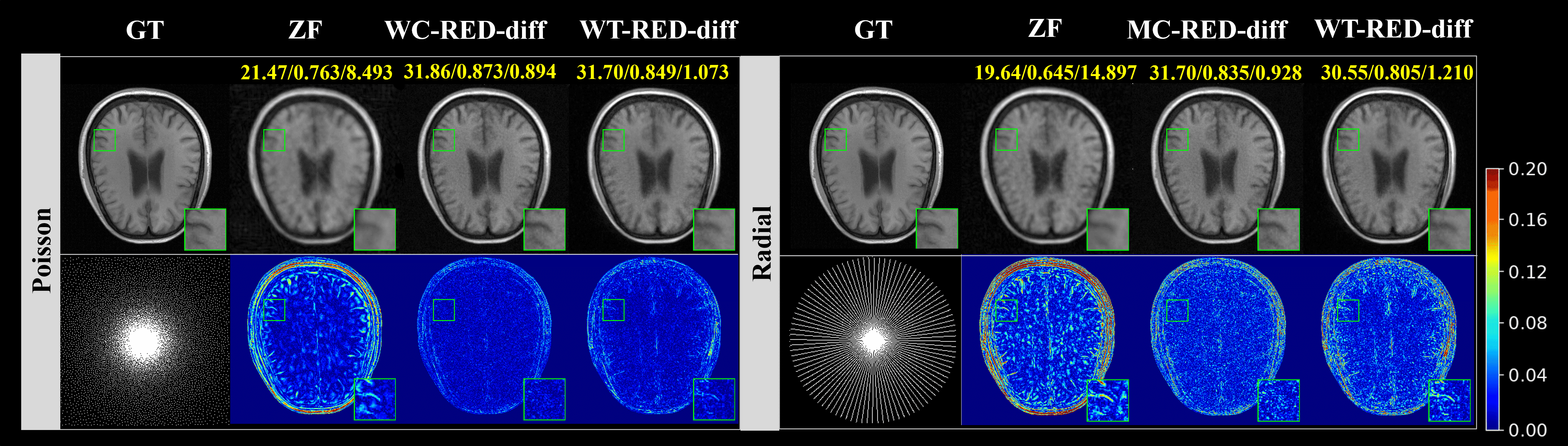}
        \captionof{figure}{Illustration of iterative reconstruction under our high-dimensional k-space formulation \(K=\mathrm{HEP}(k)\). Each subfigure depicts iterative inference of a typical diffusion reconstruction solver in the embedded space.}
        \label{fig:undersampling_red_diff}
    \end{minipage}
\end{center}
\vspace{0.6em}

\subsection{Compatibility with Low-rank Regularizers}\label{subsec:compatibility-low-rank-regularizers}

We further investigate whether the proposed HEP framework can be combined with classical structural prior. Low-rank constraints have been widely used in MRI reconstruction to exploit redundancy and structural correlation in k-space or image-domain representations. While the previous experiments focus on the standalone effectiveness of high-dimensional embedding, this study examines whether HEP can serve as a flexible representation backbone that remains compatible with additional model-based regularization. Based on the results in earlier sections, RED-diff is adopted as the base solver due to its stable performance across different settings. Experiments are conducted on the fastMRI dataset under a fixed noise level of $\mathrm{SNR}=20~\mathrm{dB}$ with Poisson sampling. To assess complementarity, we compare HEP-enhanced RED-diff with and without low-rank regularization. This comparison aims to determine whether high-dimensional representations and low-rank constraints provide synergistic benefits for reconstruction. Quantitative results are reported in Table~\ref{tab:low_rank_regularizers}.

\vspace{0.6em}
\begin{center}
\vspace{-0.5em}
\captionsetup{
    type=table,
    labelsep=newline,
    labelfont=bf
}
\captionof{table}{Reconstruction performance of HEP framework fused with low-rank prior on the fastMRI dataset (SNR = 20 dB, Poisson sampling). Metrics: PSNR (dB) / SSIM / NMSE ($10^{-2}$).}
\vspace{0.4em}
\label{tab:low_rank_regularizers}

\footnotesize
\setlength{\tabcolsep}{5pt}
\renewcommand{\arraystretch}{1.50}
\setlength{\doublerulesep}{1pt} % 调小双线间距

\begin{tabular}{>{\centering\arraybackslash}m{1cm}ccccccc}
\noalign{\hrule height 1.2pt}
\multirow[c]{2}{*}{\textbf{AF}} 
& \multicolumn{2}{c}{\textbf{{\color{heppurple}HEP-}RED-Diff}} 
& \multirow[c]{2}{*}{\textbf{LR-RED-diff}} 
& \multicolumn{2}{c}{\textbf{{\color{heppurple}HEP-}RED-Diff}} \\
\cline{2-3}\cline{5-6}
%& MC-RED-diff & WT-RED-diff 
& \textbf{Multi-channel} & \textbf{Wavelet Transform}
& 
& \textbf{Low-rank+Multi-channel} & \textbf{Low-rank+Wavelet Transform} \\
%& MC-LR-RED-diff & WT-LR-RED-diff \\
\hline
$R=6$  & 34.89/0.901/0.432 & 34.16/0.903/0.514 & 35.83/0.908/0.293 & \textbf{37.79}/\textbf{0.944}/\textbf{0.112} & 36.99/0.914/0.183 \\
$R=10$ & 33.53/0.886/0.562 & 33.38/0.868/0.578 & 34.83/0.903/0.488 & \textbf{36.54}/\textbf{0.925}/\textbf{0.163} & 35.76/0.908/0.221 \\
$R=15$ & 31.64/0.872/0.964 & 31.10/0.857/1.032 & 33.56/0.895/0.678 & \textbf{35.67}/\textbf{0.908}/\textbf{0.243} & 34.44/0.906/0.502 \\
\noalign{\hrule height 1.2pt}
\end{tabular}

\vspace{0.6em}
\end{center}
\vspace{0.6em}

As shown in Table~\ref{tab:low_rank_regularizers}, incorporating low-rank regularization further improves the reconstruction performance of HEP-based RED-diff variants in most settings. This indicates that the proposed high-dimensional representation is not limited to diffusion-prior learning, but can also be effectively combined with classical structural prior in MRI reconstruction. Compared with using low-rank regularization alone, the fusion strategy achieves superior performance, benefiting from both representation-level enrichment and explicit structural constraints. This suggests that high-dimensional embedding and low-rank modeling provide complementary regularization effects for ill-posed k-space reconstruction, leading to more robust recovery under noise and undersampling. Qualitative results in Fig.~\ref{fig:low_rank_fusion} further support these observations. The low-rank fusion variant reduces residual artifacts and improves local structural consistency, particularly in the magnified regions. Overall, these results demonstrate that HEP is compatible with conventional model-based prior and can serve as a general representation framework that accommodates additional structural regularization for more robust MRI reconstruction.

\vspace{0.9em}
\begin{center} \begin{minipage}{\linewidth} \centering \includegraphics[width=\linewidth]{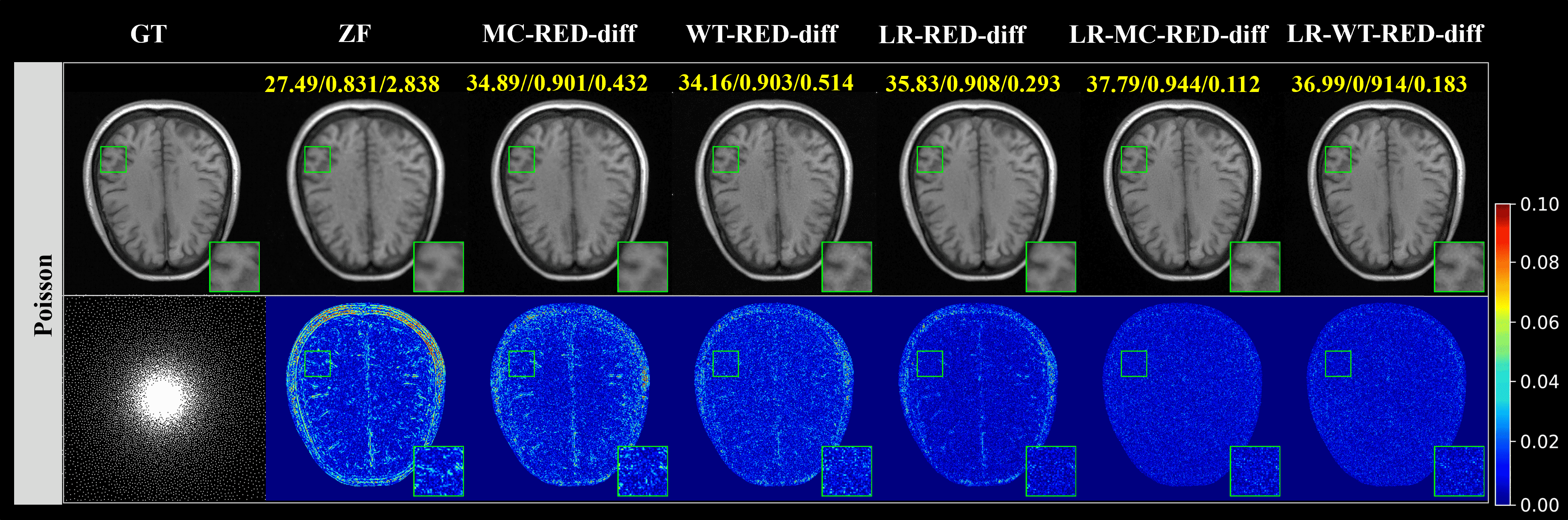} \captionof{figure}{Qualitative comparison of RED-diff variants with/without low-rank fusion on the fastMRI dataset (SNR = 20 dB, 2D Poisson sampling, $R=6$). Rows: reconstruction images, $2\times$ magnified views of red-boxed regions, residual maps.} \label{fig:low_rank_fusion} \end{minipage} \end{center}

\vspace{0.1em}
\section{Conclusion}\label{sec:conclusion}

In this work, we proposed a unified HEP framework for noisy k-space MRI reconstruction, designed to enhance diffusion-based inverse solvers from a representation perspective.By introducing a general embedding operator for high-dimensional spaces, we systematically integrated enriched k-space representations, including multi-channel and wavelet domain embeddings, into a range of diffusion solvers without modifying their core inference mechanisms. Through a unified analysis of error propagation, we demonstrated that the efficacy of these representations hinges critically on the coupling between denoising errors and data consistency across distinct algorithmic architectures, which accounts for the heterogeneous performance gains observed in practice. Experimental results further validated that embedding in high-dimensional spaces consistently enhances reconstruction quality, particularly under challenging conditions marked by aggressive undersampling and Rician noise. These findings underscore the critical role of representation design for diffusion-driven inverse problems, offer a principled framework for future advancements in k-space MRI reconstruction.

\section*{CRediT authorship contribution statement}

Yu Guan: Writing -- review \& editing, Writing -- original draft, Methodology, Conceptualization. Tianjia Huang: Writing -- review \& editing, Writing -- original draft, Methodology, Conceptualization. Qinrong Cai: Writing -- review \& editing, Resources, Data curation. Qiuyun Fan: review \& editing. Dong Liang: review \& editing. Qiegen Liu: Writing -- review \& editing, Supervision, Project administration, Funding acquisition.

\section*{Declaration of competing interest}

The authors declare that they have no known competing financial interests or personal relationships that could have appeared to influence the work reported in this paper.

\section*{Acknowledgments}

This work was supported in part by the National Key Research and Development Program of China, Grant Nos: 2023YFF1204300, 2023YFF1204302.

\bibliographystyle{cas-model2-names}

%% Loading bibliography database
\bibliography{cas-refs}

\end{document}